\documentclass[11pt]{article}

\usepackage[final]{acl}

\usepackage{times}
\usepackage{latexsym}
\usepackage{amsmath}
\usepackage{amsfonts}  
\usepackage{bm}
\usepackage{booktabs} 
\usepackage{multirow}
\usepackage{tabularx}
\usepackage{graphicx}    
\usepackage{caption}     
\usepackage{subcaption}  
\usepackage{float}
\newcolumntype{C}{>{\centering\arraybackslash}X}
\usepackage{array}     
\usepackage{makecell}  
\usepackage[most]{tcolorbox}
\tcbuselibrary{listings,breakable}
\usepackage{listings}
\usepackage[table]{xcolor}

\usepackage[T1]{fontenc}

\usepackage[utf8]{inputenc}

\usepackage{microtype}

\usepackage{inconsolata}

\usepackage{graphicx}

\title{MARS$^2$: Scaling Multi-Agent Tree Search \\via Reinforcement Learning for Code Generation}

\author{
 \textbf{Pengfei Li\textsuperscript{1,2,3}\thanks{Equal contribution}},
 \textbf{Shijie Wang\textsuperscript{1}\footnotemark[1]},
 \textbf{Fangyuan Li\textsuperscript{2,3}\footnotemark[1]},
 \textbf{Yikun Fu\textsuperscript{1}},
 \textbf{Kaifeng Liu\textsuperscript{3}},
\\
 \textbf{Kaiyan Zhang\textsuperscript{4}\thanks{Corresponding authors}},
 \textbf{Dazhi Zhang\textsuperscript{3}},
 \textbf{Yuqiang Li\textsuperscript{1,2}\footnotemark[2]},
 \textbf{Biqing Qi\textsuperscript{1}\footnotemark[2]},
 \textbf{Bowen Zhou\textsuperscript{1}},
\\
 \textsuperscript{1}Shanghai Artificial Intelligence Laboratory,
 \textsuperscript{2}Shanghai Innovation Institute,
\\
 \textsuperscript{3}Harbin Institute of Technology,
 \textsuperscript{4}Tsinghua University
 \\
 \normalsize{
   \href{mailto:lipengfei0208@gmail.com}{lipengfei0208@gmail.com}
   }
\\
\normalsize{
   \href{mailto:zhang-ky22@mails.tsinghua.edu.cn}{zhang-ky22@mails.tsinghua.edu.cn},
   \href{mailto:liyuqiang@pjlab.org.cn}{liyuqiang@pjlab.org.cn},
   \href{mailto:qibiqing7@gmail.com}{qibiqing@pjlab.org.cn}
}
}

\begin{document}
\maketitle
\begin{abstract}
Reinforcement learning (RL) paradigms have demonstrated strong performance on reasoning-intensive tasks such as code generation. However, limited trajectory diversity often leads to diminishing returns, which constrains the achievable performance ceiling. Search-enhanced RL alleviates this issue by introducing structured exploration, which remains constrained by the single-agent policy priors. 
Meanwhile, leveraging multiple interacting policies can acquire more diverse exploratory signals, but existing approaches are typically decoupled from structured search.
We propose \textbf{MARS$^2$} (Multi-Agent Reinforced Tree-Search Scaling), a unified RL framework in which multiple independently-optimized agents collaborate within a shared tree-structured search environment.
MARS$^2$ models the search tree as a learnable multi-agent interaction environment, enabling heterogeneous agents to collaboratively generate and refine candidate solutions within a shared search topology. To support effective learning, we introduce a path-level group advantage formulation based on tree-consistent reward shaping, which facilitates effective credit assignment across complex search trajectories.
Experiments on code generation benchmarks show that MARS$^2$ consistently improves performance across diverse model combinations and training settings, demonstrating the effectiveness of coupling multi-agent collaboration with tree search for enhancing reinforcement learning. 
Our code is publicly available at \url{https://github.com/TsinghuaC3I/MARTI}.
\end{abstract}

\section{Introduction}
\begin{figure*}[t]
    \centering
    \includegraphics[width=1.0\textwidth]{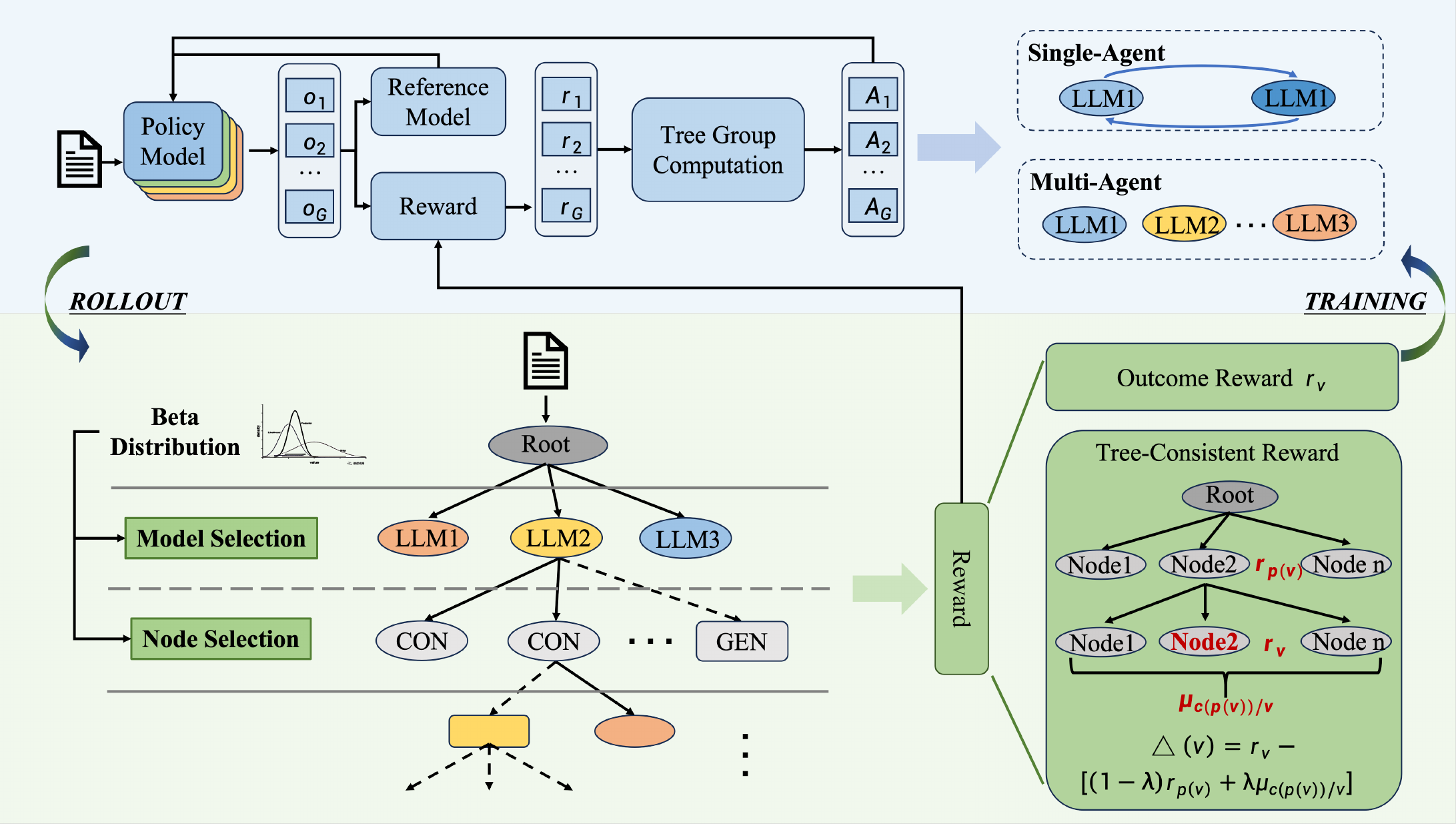}
    \caption{\textbf{Overview of the MARS$^{2}$ framework.} Multiple agents collaboratively expand a shared search tree via Thompson-sampling-based agent--node selection, with node-level rewards refined by tree-consistent reward shaping over parent and sibling signals. Each agent is then independently optimized with a tree-level group-relative advantage over the shared tree.}
    \vspace{-5pt}
    \label{fig:framework}
\end{figure*}
In recent years, reinforcement learning (RL) paradigms represented by Group Relative Policy Optimization (GRPO)\citep{shao2024deepseekmath} have achieved notable progress on reasoning-intensive tasks such as mathematical problem solving and code generation. By directly optimizing policies with respect to outcome-level feedback, these methods substantially improve response quality under single-sample evaluation\citep{areal, deepcoder2025}. However, such improvements are still largely constrained by the exploration behavior of a single agent\citep{marti2025}. Under the independent and identically distributed (i.i.d.) sampling assumption, the effective exploration space is implicitly bounded by the model’s own prior distribution, which often leads to insufficient trajectory diversity and premature convergence to local optima, thereby imposing a hard-to-break performance ceiling\citep{song2025outcomebased, zhang2025survey}.

To alleviate this exploration bottleneck, search-augmented reinforcement learning introduces explicit structured exploration mechanisms. Representative approaches such as TreeRL\citep{treerl} integrate Monte Carlo Tree Search (MCTS) into the training process, explicitly expanding the space of candidate solutions under a fixed computational budget. Prior studies\citep{treerl, zhoubian2025restrlachievingaccuratecode, li2025treepo} have shown that tree-structured search can improve trajectory diversity and provide higher-quality optimization signals. Nevertheless, in most existing methods\citep{treerl, zhoubian2025restrlachievingaccuratecode, li2025treepo}, the entire search tree is still driven by a single policy distribution, and the resulting search dynamics remain fundamentally constrained by a shared prior. As training progresses, search behavior increasingly concentrates on a small number of high-probability branches, making it difficult to continuously expand the exploration frontier and leading to diminishing returns from search \citep{silver2016alphago} (\textbf{Challenge 1: diminishing exploration gains under a single-policy prior}).

From a complementary perspective, multi-agent reinforcement learning (MARL) is widely regarded as a promising approach to overcoming the exploration limitations of single-policy optimization\citep{DBLP:journals/corr/Hernandez-LealK17}. By introducing interactions among multiple policies, MARL induces non-stationary data distributions that evolve with agent interactions, thereby offering the potential to escape the exploration bounds imposed by a single policy\citep{DBLP:conf/nips/LoweWTHAM17}. Recent studies in large language model (LLM) reasoning settings, such as MAPoRL\citep{DBLP:conf/acl/Park0GOZK25}, further demonstrate that incorporating multi-agent collaboration into the training loop can yield more stable and accumulative performance gains than relying on multi-agent protocols solely at inference time\citep{inoue2025wider}. 
However, existing multi-agent reasoning frameworks predominantly rely on simple interaction paradigms, such as multi-round dialogue, debate, or voting~\citep{DBLP:conf/acl/ZhangX25, zuo2025ttrl}. These methods treat agent interaction as a lightweight coordination mechanism rather than a structured exploration process. As a result, multi-agent collaboration remains largely disconnected from the underlying search dynamics, lacking explicit support for branching, backtracking, or principled allocation of exploration budget. This limitation prevents agent interactions from effectively shaping the exploration frontier and constrains their applicability to deep, multi-branch reasoning tasks (\textbf{Challenge 2: lack of structured search integration in multi-agent collaboration}).


To address these two challenges, we propose \textbf{MARS$^2$} (Multi-Agent Reinforced Tree-Search Scaling), a unified reinforcement learning framework in which multiple independently-optimized policies collaborate within a shared tree-structured search environment (Figure \ref{fig:framework}).
Unlike prior approaches that treat search as a one-off sampling procedure, MARS$^2$ models the search tree as a structured and learnable multi-agent environment. Multiple agents collaboratively explore a shared search tree, and effective credit assignment is achieved through a path-level group advantage function together with tree-consistent reward shaping. This design enables stable reward allocation across complex search trajectories and mitigates the non-stationarity induced by multiple policies concurrently updating over a shared search tree.
We conduct extensive evaluations on code generation benchmarks, where experimental results demonstrate that MARS$^2$ consistently improves performance across diverse model combinations and training settings, while exhibiting clear advantages on system-level search metrics. These findings suggest that integrating multi-agent collaboration with structured tree search provides an effective pathway for overcoming single-policy exploration bottlenecks in search-augmented reinforcement learning.

The main contributions of this work are summarized as follows:
\begin{itemize}
    \item \textbf{A unified multi-agent tree search augmented reinforcement learning framework.}
    We introduce MARS$^2$, which embeds multi-agent collaboration directly into structured tree search, transforming search from a static sampling procedure into a learnable, cooperative exploration process.

    \item \textbf{Path level group advantage modeling and reward allocation.}  
    We propose a group-level advantage formulation defined over search paths, enabling stable credit assignment and reward shaping under tree-structured constraints, thereby improving training stability in MARS$^2$.

    \item \textbf{Systematic empirical evaluation and ablation studies.}  
    We conduct comprehensive experiments on code generation benchmarks across multiple model combinations and settings, analyzing the effects of multi-agent collaboration, structured search, and reward mechanisms, and demonstrating the robustness and effectiveness of the proposed approach.
\end{itemize}

\section{Methodology}
\label{sec:methodology}

\subsection{Problem Formulation and Preliminary}

We consider complex reasoning tasks such as code generation, and model the inference process as a sequential decision-making problem under a finite computational budget. Given an input problem $x \in \mathcal{X}$ and a reasoning budget $N$, the standard parallel sampling paradigm independently draws a set of candidate solutions from the model distribution:
$$
\mathbf{y} = \{y_1, y_2, \dots, y_N\}, \quad y_i \stackrel{\text{i.i.d.}}{\sim} f_M(\cdot \mid x),
$$
where $f_M(\cdot)$ denotes the output distribution parameterized by model $M$.

In contrast to parallel sampling, we model the reasoning process as a multi-agent collaborative tree search, and treat the search tree itself as a learnable interaction environment. Specifically, multiple agents collaboratively expand a shared search tree, producing a set of tree nodes
$$
\hat{\mathbf{y}} = \{\hat{y}_1, \hat{y}_2, \dots, \hat{y}_N\},
$$
where each $\hat{y}_i$ corresponds to a complete candidate solution generated at the $i$-th expansion step.


Following the AB-MCTS formulation~\citep{inoue2025wider}, we introduce explicit agent selection and node expansion mechanisms. 
At each expansion step, the search is formulated as a multi-armed bandit problem over agent--node pairs. To this end, we maintain Beta priors to model the selection probabilities over agents and their associated expandable nodes. Thompson sampling\citep{Thompson1933ONTL} is first applied to select the most promising agent, and is then applied again to choose a node associated with the selected agent for expansion.
During node selection, we distinguish two functional types of expandable nodes. 
A generation node is a virtual option corresponding to proposing a new candidate solution by the selected agent, whereas a refinement node refers to an existing tree node previously generated by that agent along the current search path. 
If a generation node is selected, the search performs a horizontal expansion by generating a new candidate solution; otherwise, it proceeds with a vertical refinement by selecting an existing child node of the current node. This design enables a dynamic balance between exploiting high-quality trajectories and exploring diverse candidate solutions.

Under this framework, the generation of each tree node $\hat{y}_i$ can be formalized as a joint distribution:
$$
\hat{y}_i \sim g_i(m)\cdot f(\cdot \mid x, m, \hat{y}_{\tau(i)}, o),
$$
where $g_i(m)$ denotes the probability of selecting agent $m \in \{M_1, M_2, \dots, M_m\}$ at the $i$-th expansion, $\hat{y}_{\tau(i)}$ is the parent node, and $o$ represents observable environment feedback (e.g., execution results or error messages).

\subsection{Tree-Node-Level Reward Assignment}

In reinforcement learning for reasoning tasks, GRPO\citep{shao2024deepseekmath} distinguishes rollout quality using group-relative advantages, which has been shown to provide stable and effective optimization signals. Parallel sampling naturally forms a group of rollouts with rewards
$
\mathbf{r} = \{r_1, r_2, \dots, r_N\},
$
from which the normalized group advantage is computed as
\begin{equation}
\label{equ:advantage}
A_i = \frac{r_i - \mathrm{mean}(\mathbf{r})}{\mathrm{std}(\mathbf{r})}.
\end{equation}

In multi-agent tree search, all nodes in a search tree originate from the same input problem, and therefore naturally constitute a semantic group. We thus extend the group-relative advantage formulation \ref{equ:advantage} to the tree-level, and define the tree-level relative advantage as
\begin{equation}
\label{equ:tree advantage}
\hat{A}_{k,j_k} =
\frac{
r_{k,j_k} - \mathrm{mean}(r_{1,j_1}, r_{2,j_2}, \dots, r_{N,j_N})
}{
\mathrm{std}(r_{1,j_1}, r_{2,j_2}, \dots, r_{N,j_N})
},
\end{equation}
where $r_{k,j_k}$ denotes the reward of the node generated at the $k$-th expansion by agent $j_k$.

However, nodes in a multi-agent tree search are not generated independently. Instead, agents iteratively refine or diversify solutions through structured interactions along the tree. A purely global tree-level advantage fails to capture such hierarchical dependencies. Intuitively, a child node should not only achieve a high global reward, but also improve upon its parent and outperform sibling candidates under the same parent.

\subsection{Tree-Consistent Reward Shaping}

To explicitly model hierarchical collaboration, we introduce a tree-consistent reward shaping mechanism. For any non-root node $v$, we define a mixed baseline
\begin{equation}
\label{equ:reward baseline}
    b(v) = (1 - \lambda)\, r_{p(v)} + \lambda \cdot \mu_{C(p(v)) \setminus v},
\end{equation}
where $r_{p(v)}$ is the reward of the parent node, $\mu_{C(p(v)) \setminus v}$ denotes the mean reward of sibling nodes sharing the same parent, and $\lambda \in [0,1]$ balances vertical improvement and horizontal competition. When a parent node produces only a single child, the baseline naturally degenerates to the parent reward.

We then define the structural consistency gain as
\begin{equation}
\label{equ:reward delta}
\Delta(v) = r_v - b(v),
\end{equation}
and apply reward shaping as
\begin{equation}
\label{equ:reward shaping}
\hat{r}_v = r_v + \gamma \cdot \Delta(v),
\end{equation}
where $\gamma$ controls the shaping strength. This mechanism preserves the original task reward semantics while explicitly encouraging both parent-child improvement and sibling-level competition, thereby facilitating effective collaboration and specialization among agents within the shared search tree. 
The above formulation relies only on a node-level scalar reward $r_v$, and is therefore independent of the specific reward source.

\subsection{Optimization Objective}

During training, we adopt GRPO-style policy optimization as the base framework. The GRPO objective can be written as
\begin{equation}
\label{equ:grpo}
\begin{aligned}
&\mathcal{J}_{\mathrm{GRPO}}(\theta)
 \\
=&\mathbb{E}_{q \sim P(Q),\, \{o_i\}_{i=1}^N \sim \pi_{\theta_{\mathrm{old}}}(\cdot \mid q)}\Bigg[
\frac{1}{N} \sum_{i=1}^N \frac{1}{|o_i|}
\sum_{t=1}^{|o_i|}
\\
& \Big(
\min\!\left(
w(i,t)\, A_i,\,
\mathrm{clip}\!\left(w(i,t), 1-\epsilon, 1+\epsilon\right) A_i
\right)
\\
& - \beta\, \mathbb{D}_{\mathrm{KL}}\!\left(\pi_\theta \,\|\, \pi_{\mathrm{ref}}\right)
\Big)
\Bigg],
\end{aligned}
\end{equation}
where $A_i$ denotes the group-relative advantage and
\[
w(i,t) = \frac{\pi_\theta(o_{i,t} \mid q, o_{i,<t})}{\pi_{\theta_{\mathrm{old}}}(o_{i,t} \mid q, o_{i,<t})}.
\]

Building on multi-agent tree search and tree-consistent reward shaping, we extend the optimization unit from parallel trajectories to tree nodes, yielding the final objective, where each agent is optimized independently using the rewards collected at the tree nodes it generates:
\begin{equation}
\label{equ:mars}
\begin{aligned}
&\mathcal{J}_{\mathrm{MARS}^2}(\bm{\Theta})
\\
=&\mathbb{E}_{q \sim \mathcal{D},\, \{o_i\}_{i=1}^N \sim \pi_{\bm{\Theta}_{\mathrm{old}}}(\cdot \mid q)}\Bigg[
\frac{1}{N} \sum_{j=1}^m
\sum_{v \in \mathcal{T}_j(q)}
\sum_{t=1}^{|o_v|}
\\
&\Big(
\min\big(
w(v,t,\theta_j)\hat{A}_{v,j},\,
\\
& \mathrm{clip}(w(v,t,\theta_j), 1-\epsilon_{\mathrm{low}}, 1+\epsilon_{\mathrm{high}})\hat{A}_{v,j}
\big) 
\\
& - \beta\, \mathbb{D}_{\mathrm{KL}}(\pi_{\theta_j} \| \pi_{\mathrm{ref}})
\Big)
\Bigg],
\end{aligned}
\end{equation}
where
\[
w(v,t,\theta_j) =
\frac{\pi_{\theta_j}(o_{v,t} \mid q, o_{v,<t})}
{\pi_{\theta_{j,\mathrm{old}}}(o_{v,t} \mid q, o_{v,<t})},
\]
$\bm{\Theta} = \{\theta_1, \theta_2, \dots, \theta_m\}$ denotes the set of agent parameters, $m$ is the number of agents, and $\mathcal{T}_j(q)$ denotes the node expand index set of the $j$-th agent. The advantage estimator $\hat{A}_{v,j}$ in Eq.~(\ref{equ:mars}) combines the tree-level group advantage of Eq.~(\ref{equ:tree advantage}) with the reward shaping of Eq.~(\ref{equ:reward shaping}):
\begin{equation}
\label{equ:advantage and reward shaping}
\hat{A}_{v,j} = \frac{\hat{r}_{v,j} - \mathrm{mean}(\hat{r}_{1,j_1}, \ldots, \hat{r}_{N,j_N})}{\mathrm{std}(\hat{r}_{1,j_1}, \ldots, \hat{r}_{N,j_N})},
\end{equation}
where $\hat{r}_{v,j}$ denotes the shaped reward of node $v$ generated by agent $j$, and $N$ is the number of nodes expanded during tree search. Agent--node selection within $\mathcal{T}_j(q)$ adopts standard Thompson sampling~\citep{Thompson1933ONTL} and is orthogonal to our core contribution: the tree-level group-relative advantage and its hierarchical credit assignment.

\section{Experiments}
\label{sec:experiments}

\subsection{Experimental Setup}
\paragraph{Datasets.}
We use the open-source code generation dataset released by DeepCoder\citep{deepcoder2025} as the training data for reinforcement learning.
To improve training efficiency and avoid degenerate reward signals, we filter out two types of samples following Polaris\cite{Polaris2025}:
(i) prompts for which the model consistently achieves a perfect score, providing little optimization signal, and
(ii) prompts for which all sampled solutions receive zero reward, resulting in overly sparse and noisy gradients.
After filtering, the resulting training set contains 7,992 code generation prompts.
All reinforcement learning experiments, including MARS$^2$ and all baselines, are conducted on this dataset.

\paragraph{Models.}
To evaluate the effectiveness of MARS$^2$ across different model types and scales, we consider open-source large language models with parameter sizes of 8B and 14B, covering both code-oriented and general-purpose models.
Specifically, we adopt \textbf{AReaL-boba-2 8B/14B}\citep{areal} and \textbf{DeepCoder-14B-Preview}\citep{deepcoder2025} as code-specialized models that have been extensively optimized for programming tasks.
In addition, we include \textbf{Qwen3 8B/14B}\citep{yang2025qwen3} to assess the generality of MARS$^2$ beyond code-centric pretraining.

\paragraph{Baselines.}
At the algorithmic level, we adopt Vanilla GRPO as the single-agent reinforcement learning baseline.
To ensure a fair comparison and eliminate the influence of additional reinforcement learning tricks, we incorporate the same training techniques used in MARS$^2$ into the Vanilla GRPO baseline (Appendix~\ref{sec:appendix-Training Details} \& Appendix~\ref{sec:appendix-training-strategy}).
We further verify in Appendix~\ref{sec:appendix-Additional Performance} (Table~\ref{tab:stabilization}) that these stabilization techniques alone do not account for the reported gains, as GRPO equipped with them still remains substantially below RS$^2$ and MARS$^2$.
In addition, all tree search methods share the same prompt templates and formatting conventions, as detailed in Appendix~\ref{sec:appendix-Prompt Template}.
As a result, performance differences can be primarily attributed to multi-agent coordination and structured search, rather than optimization details.

\paragraph{Training and Inference Configurations.}
We explicitly distinguish how training and inference are configured across all methods:
\begin{itemize}
    \item \textbf{Training.} Vanilla GRPO follows the standard parallel sampling scheme of Eq.~(\ref{equ:grpo}) without tree-structured procedure, whereas RS$^2$ and MARS$^2$ expand $n$ tree nodes per prompt with each $o_i$ corresponding to one tree node. For fairness, every agent across all methods is trained on the same number of samples and optimization steps.
    \item \textbf{Inference.} All methods, including those trained under GRPO, are evaluated under a unified MCTS-based inference framework with a fixed search budget of 60 nodes for Pass@1 (MCTS) and Pass@N.
\end{itemize}
These two phases give rise to two orthogonal notions of \emph{single-} vs \emph{multi-agent}: a model can be trained with a single policy (GRPO, RS$^2$) or multiple policies over a shared tree (MARS$^2$), and at inference time evaluated either via standard single-pass sampling (Pass@1) or via MCTS-based search (Pass@1 (MCTS), Pass@N). The two dimensions are evaluated independently, allowing us to disentangle the contributions of training and inference mechanisms. Throughout the paper, we use "(Q+A)" marks a single model jointly trained on Qwen3 and AReaL; "\&" marks a multi-model inference system.

\definecolor{marsblue}{RGB}{220,235,252}
\definecolor{gainred}{RGB}{200,30,30}
\definecolor{lossgreen}{RGB}{30,130,60}
\vspace{-5pt}
\begin{table*}[t]
\centering
\caption{\textbf{Overall performance across models and systems under different training methods.}
Single-model performance is measured by Pass@1, while system-level performance is evaluated using Pass@1 (MCTS) and Pass@N under a fixed inference budget of 60.
Results compare Base, GRPO, RS$^2$, and MARS$^2$ across both individual models and multi-model systems.
\textbf{Notation:} ``Q'' and ``A'' abbreviate Qwen3 and AReaL respectively; ``+'' (e.g., \mbox{Q+A}) denotes a model combination used at training time, while ``\&'' (e.g., Q\&A) denotes a multi-model combination used at inference time.}
\label{tab:main_results}
\small
\setlength{\tabcolsep}{18pt}
\renewcommand{\arraystretch}{0.9}
\begin{tabular}{lllll}
\toprule
\textbf{Model / System} & \textbf{Method} & \textbf{Pass@1} & \textbf{Pass@1 (MCTS)} & \textbf{Pass@N} \\
\midrule
\multirow{4}{*}{Qwen3-8B}
& Base            & 50.3 & 54.3 & 68.6 \\
& GRPO            & 52.5$_{\color{gainred}+2.2}$ & 57.1$_{\color{gainred}+2.8}$ & \textbf{73.1}$_{\color{gainred}\boldsymbol{+4.5}}$ \\
& RS$^2$          & \underline{55.4}$_{\color{gainred}\underline{+5.1}}$ & \underline{60.6}$_{\color{gainred}\underline{+6.3}}$ & 71.4$_{\color{gainred}+2.8}$ \\
\rowcolor{marsblue} \cellcolor{white} & MARS$^2$ (Q+A) & \textbf{58.3}$_{\color{gainred}\boldsymbol{+8.0}}$ & \textbf{60.8}$_{\color{gainred}\boldsymbol{+6.5}}$ & \underline{72.3}$_{\color{gainred}\underline{+3.7}}$ \\
\midrule
\multirow{4}{*}{AReaL-8B}
& Base            & 51.0 & 56.0 & 66.9 \\
& GRPO            & 50.8$_{\color{lossgreen}-0.2}$ & 55.1$_{\color{lossgreen}-0.9}$ & 67.4$_{\color{gainred}+0.5}$ \\
& RS$^2$          & \textbf{55.4}$_{\color{gainred}\boldsymbol{+4.4}}$ & \textbf{58.3}$_{\color{gainred}\boldsymbol{+2.3}}$ & \underline{69.7}$_{\color{gainred}\underline{+2.8}}$ \\
\rowcolor{marsblue} \cellcolor{white} & MARS$^2$ (Q+A) & \underline{54.9}$_{\color{gainred}\underline{+3.9}}$ & \underline{58.1}$_{\color{gainred}\underline{+2.1}}$ & \textbf{70.3}$_{\color{gainred}\boldsymbol{+3.4}}$ \\
\midrule
\multirow{4}{*}{Qwen3-14B}
& Base            & 56.0 & 61.7 & 75.4 \\
& GRPO            & 58.3$_{\color{gainred}+2.3}$ & 62.1$_{\color{gainred}+0.4}$ & 76.6$_{\color{gainred}+1.2}$ \\
& RS$^2$          & \underline{61.1}$_{\color{gainred}\underline{+5.1}}$ & \underline{65.1}$_{\color{gainred}\underline{+3.4}}$ & \underline{78.9}$_{\color{gainred}\underline{+3.5}}$ \\
\rowcolor{marsblue} \cellcolor{white} & MARS$^2$ (Q+A) & \textbf{61.7}$_{\color{gainred}\boldsymbol{+5.7}}$ & \textbf{66.2}$_{\color{gainred}\boldsymbol{+4.5}}$ & \textbf{78.9}$_{\color{gainred}\boldsymbol{+3.5}}$ \\
\midrule
\multirow{4}{*}{AReaL-14B}
& Base            & 58.4 & 62.9 & 74.3 \\
& GRPO            & 58.9$_{\color{gainred}+0.5}$ & 60.7$_{\color{lossgreen}-2.2}$ & 75.4$_{\color{gainred}+1.1}$ \\
& RS$^2$          & \underline{62.3}$_{\color{gainred}\underline{+3.9}}$ & \underline{68.0}$_{\color{gainred}\underline{+5.1}}$ & \textbf{81.1}$_{\color{gainred}\boldsymbol{+6.8}}$ \\
\rowcolor{marsblue} \cellcolor{white} & MARS$^2$ (Q+A) & \textbf{64.6}$_{\color{gainred}\boldsymbol{+6.2}}$ & \textbf{68.1}$_{\color{gainred}\boldsymbol{+5.2}}$ & \underline{80.2}$_{\color{gainred}\underline{+5.9}}$ \\
\midrule
\multirow{4}{*}{Q\&A-8B}
& Base            & -- & 57.2 & 69.7 \\
& GRPO            & -- & 56.0$_{\color{lossgreen}-1.2}$ & 72.0$_{\color{gainred}+2.3}$ \\
& RS$^2$          & -- & \underline{57.2}$_{\underline{+0.0}}$ & \underline{72.6}$_{\color{gainred}\underline{+2.9}}$ \\
\rowcolor{marsblue} \cellcolor{white} & MARS$^2$ (Q+A) & -- & \textbf{61.7}$_{\color{gainred}\boldsymbol{+4.5}}$ & \textbf{75.4}$_{\color{gainred}\boldsymbol{+5.7}}$ \\
\midrule
\multirow{4}{*}{Q\&A-14B}
& Base            & -- & 62.9 & 74.9 \\
& GRPO            & -- & 61.7$_{\color{lossgreen}-1.2}$ & 78.9$_{\color{gainred}+4.0}$ \\
& RS$^2$          & -- & \underline{68.0}$_{\color{gainred}\underline{+5.1}}$ & \underline{79.4}$_{\color{gainred}\underline{+4.5}}$ \\
\rowcolor{marsblue} \cellcolor{white} & MARS$^2$ (Q+A) & -- & \textbf{68.9}$_{\color{gainred}\boldsymbol{+6.0}}$ & \textbf{79.4}$_{\color{gainred}\boldsymbol{+4.5}}$ \\
\bottomrule
\end{tabular}
\vspace{-6pt}
\end{table*}

\paragraph{Evaluation Benchmark and Metrics.}

We evaluate all methods on \textbf{LiveCodeBench (v6, 01/2025–05/2025)}~\citep{livecodebench}, which was released after the development of all base models, thereby minimizing the risk of data contamination and enabling a reliable assessment of generalization. 
For each problem, the training reward is derived from passing the full set of training test cases, while public test cases guide MCTS expansion at inference and private test cases measure final performance to prevent data leakage. 
We report three complementary metrics to evaluate MARS$^2$ from different perspectives:
\begin{enumerate}
    \item \textbf{Pass@1.}
    This metric measures the probability that a model generates a correct solution in a single sampling attempt, reflecting the intrinsic code generation and reasoning capability of the learned policy.


    \item \textbf{Pass@1(MCTS).}
    During inference, we perform $N$ complete rollouts in the structured environment defined by MARS$^2$ and select as the final output the response that passes the public tests and is generated at the latest expansion step. This "latest-wins" rule is consistent with the refinement dynamics of tree search, as deeper nodes typically correspond to candidates refined along higher-quality branches. We empirically verify its advantage over random selection in Appendix~\ref{sec:appendix-Additional Performance} (Table~\ref{tab:latest-wins}).


    \item \textbf{Pass@N.}
    This metric measures the probability that at least one out of $N$ generated solutions passes all unit tests, reflecting both average performance and solution diversity.
\end{enumerate}


\subsection{Overall Performance}
\begin{figure*}[t]
    \centering

    \begin{subfigure}[b]{0.24\textwidth}
        \centering
        \includegraphics[width=\linewidth]{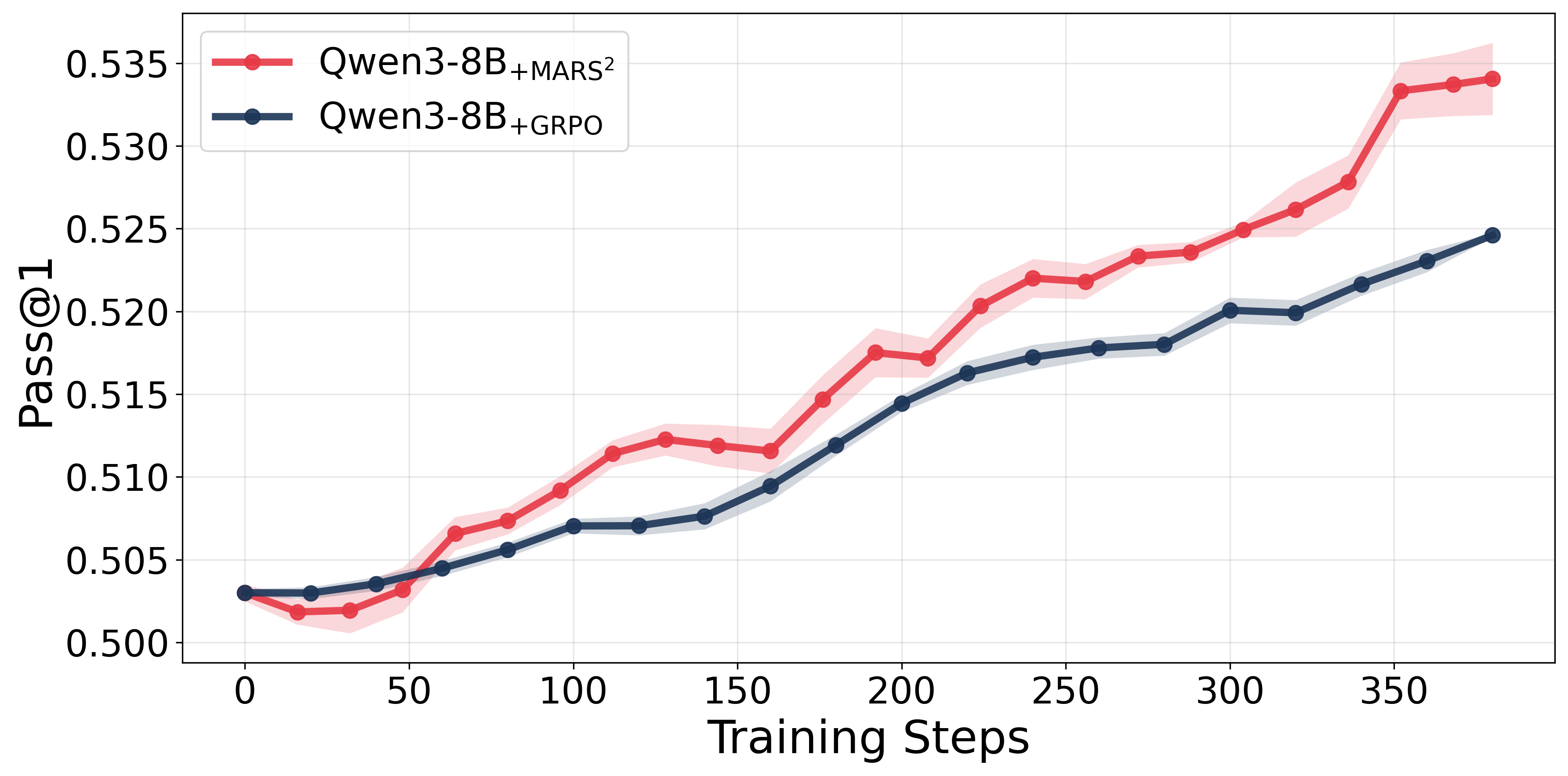}
        \subcaption{Qwen3-8B.}
        \label{fig:Qwen 8B}
    \end{subfigure}
    \hfill
    \begin{subfigure}[b]{0.24\textwidth}
        \centering
        \includegraphics[width=\linewidth]{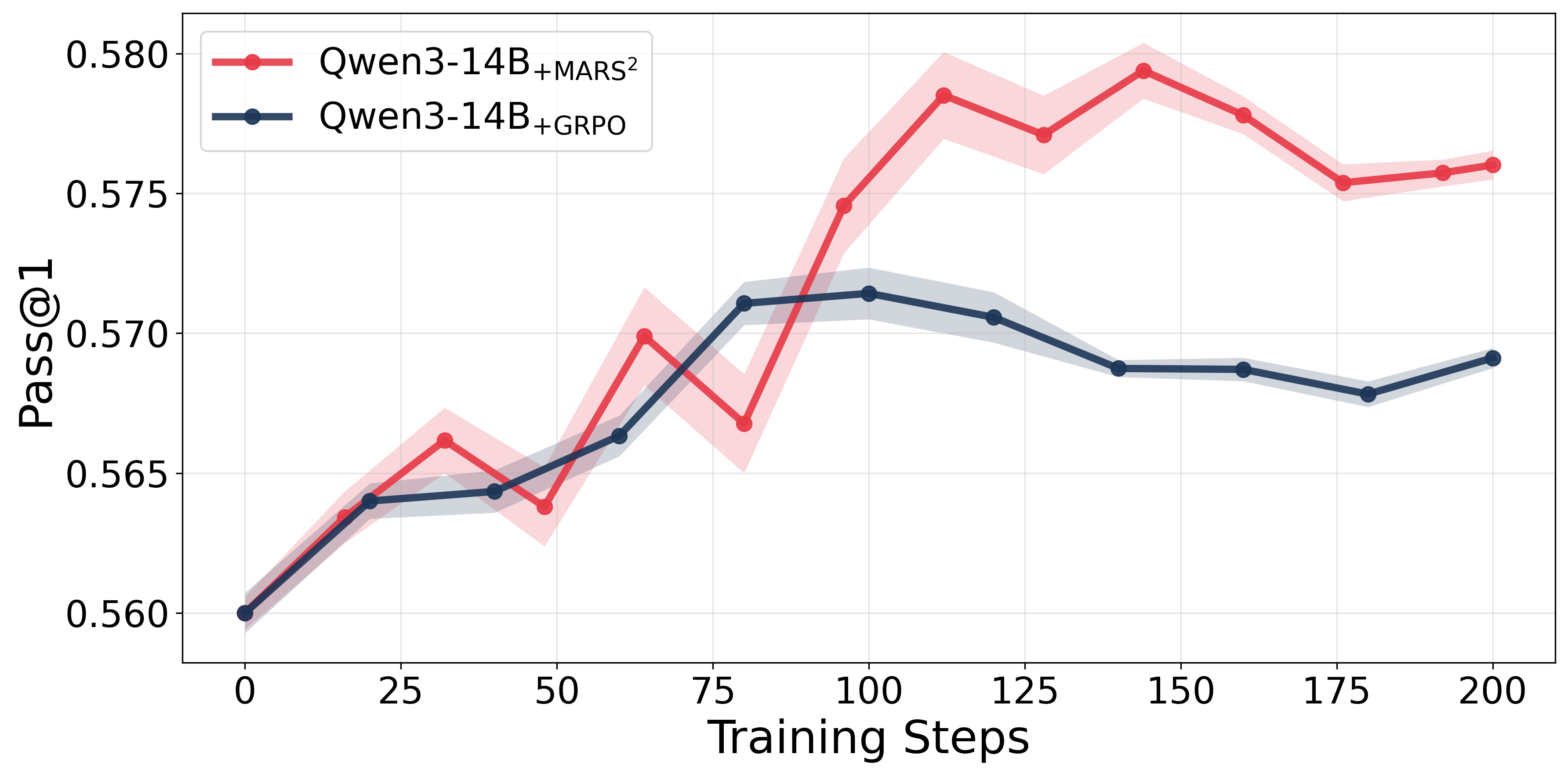}
        \subcaption{Qwen3-14B.}
        \label{fig:Qwen 14B}
    \end{subfigure}
    \hfill
    \begin{subfigure}[b]{0.24\textwidth}
        \centering
        \includegraphics[width=\linewidth]{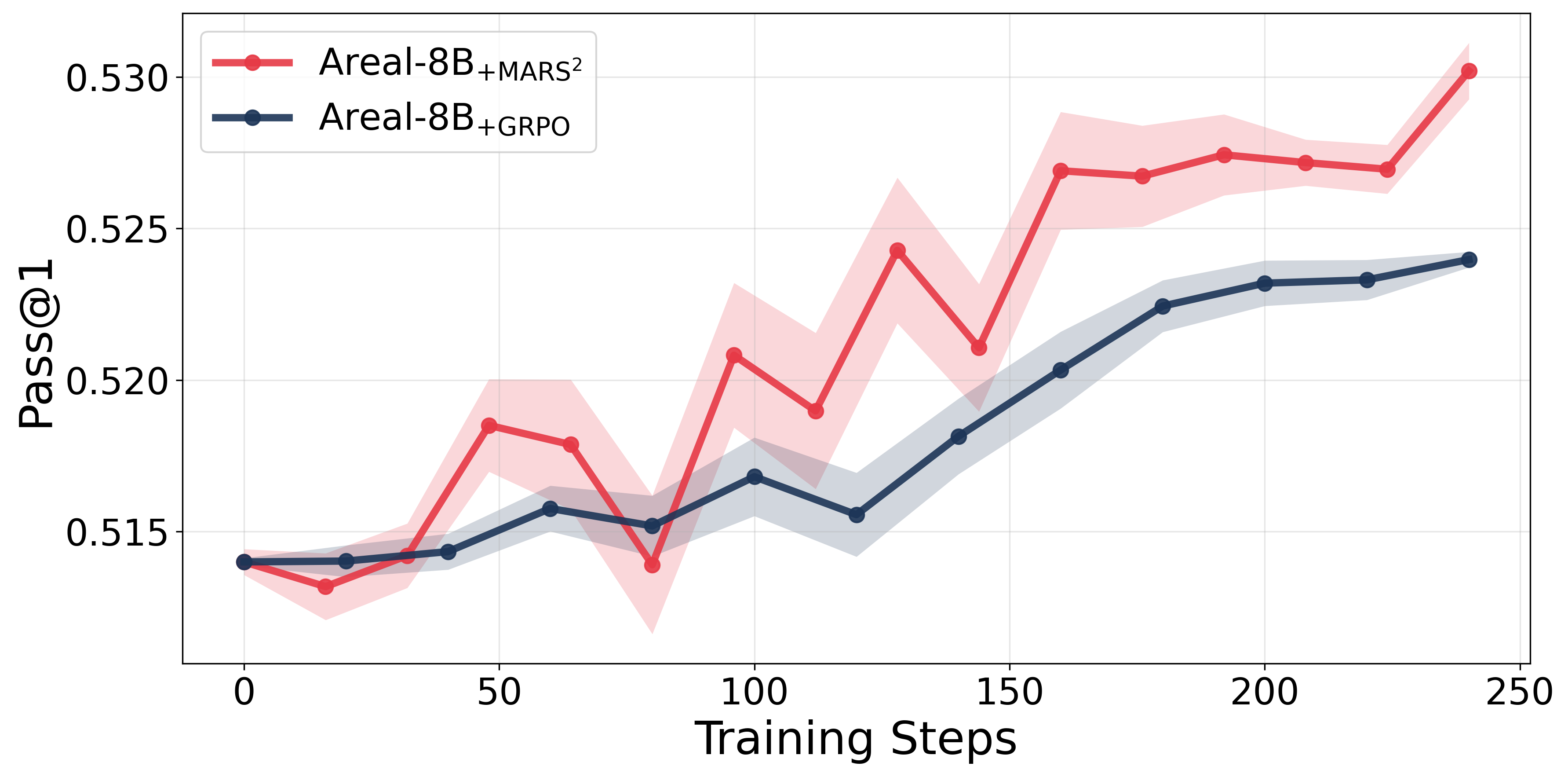}
        \subcaption{AReaL-8B.}
        \label{fig:AReaL 8B}
    \end{subfigure}
    \hfill
    \begin{subfigure}[b]{0.24\textwidth}
        \centering
        \includegraphics[width=\linewidth]{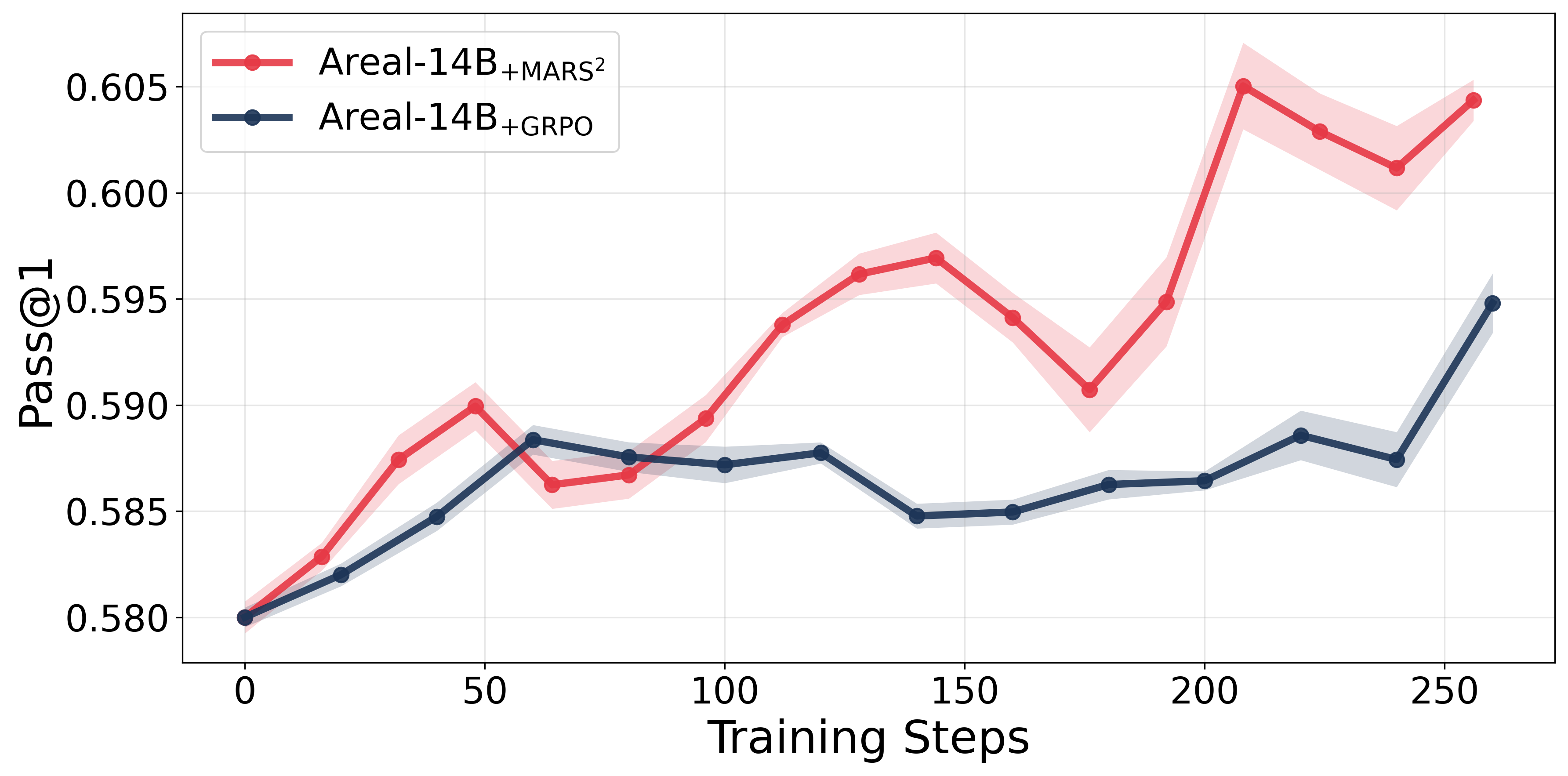}
        \subcaption{AReaL-14B.}
        \label{fig:AReaL 14B}
    \end{subfigure}
    
    \vspace{-5pt}
    \caption{\textbf{Pass@1 accuracy over training steps for multi-agent MARS$^2$ training on paired models.} Results are shown for Qwen-8B + AReaL-8B and Qwen-14B + AReaL-14B.}
    \vspace{-5pt}
    \label{fig:matraincurve}
\end{figure*}


We first evaluate the overall effectiveness of introducing tree-structured search into policy training, and then examine whether this benefit can be further extended by introducing multiple collaborating policies into the shared tree search.
Throughout this section, we intentionally exclude reward shaping and other stabilization techniques in order to isolate the contribution of the search structure itself.
\subsubsection{Single-Agent Reinforced Tree Search}
\paragraph{Effectiveness of Tree-Structured Search in Single-Agent Training.}
To evaluate whether incorporating a tree-structured search alone can yield tangible benefits for policy optimization, we first conduct systematic experiments under a single-agent setting(\textbf{RS$^2$}). Our evaluation covers five base LLMs with parameter scales of 8B and 14B. For fair comparison, all methods are trained and evaluated under identical training steps and data budgets. 
Notably, RS$^2$ does not consume more rollouts or generate more tokens than GRPO during training (see Appendix~\ref{sec:appendix-Additional Performance}, Table~\ref{tab:compute-stats}).
The overall results are summarized in Table~\ref{tab:main_results}.
Experimental results show that RS$^2$ consistently outperforms Vanilla GRPO across all base models. In terms of single-model capability, Qwen3-8B achieves the largest absolute Pass@1 improvement of 5.1\% under RS$^2$, whereas Vanilla GRPO yields only a 2.2\% gain. Moreover, we verify in Appendix~\ref{sec:appendix-Additional Performance} (Table~\ref{tab:budget-scaling}) that RS$^2$ scales gracefully with the training-time search budget: as the number of tree nodes increases from 4 to 16, we observe consistent gains followed by stable performance, with no degradation or training instability.

At the system level, RS$^2$ improves Pass@1 (MCTS) by 6.3\% relative to the base model, compared to a 2.8\% improvement from Vanilla GRPO, indicating that policies trained with RS$^2$ are more effective at guiding MCTS-based search. 
Notably, for models already highly optimized for code generation (e.g., AReaL), Vanilla GRPO offers little room for further improvement and may even degrade performance. For instance, the Pass@1 score of AReaL-8B decreases from 51.0\% to 50.8\% under Vanilla GRPO.
While RS$^2$ alleviates exploration inefficiency by exposing the policy to higher-quality search-guided trajectories, it remains driven by a single policy prior and thus cannot fundamentally expand the exploration frontier; as training proceeds, the search distribution may concentrate on a small set of high-probability branches, leading to diminishing gains and occasional instability. We provide a more detailed analysis of these limitations in Appendix~\ref{sec:appendix-Additional Performance}. 

\begin{figure*}[t]
  \centering
  \includegraphics[width=1.0\textwidth]{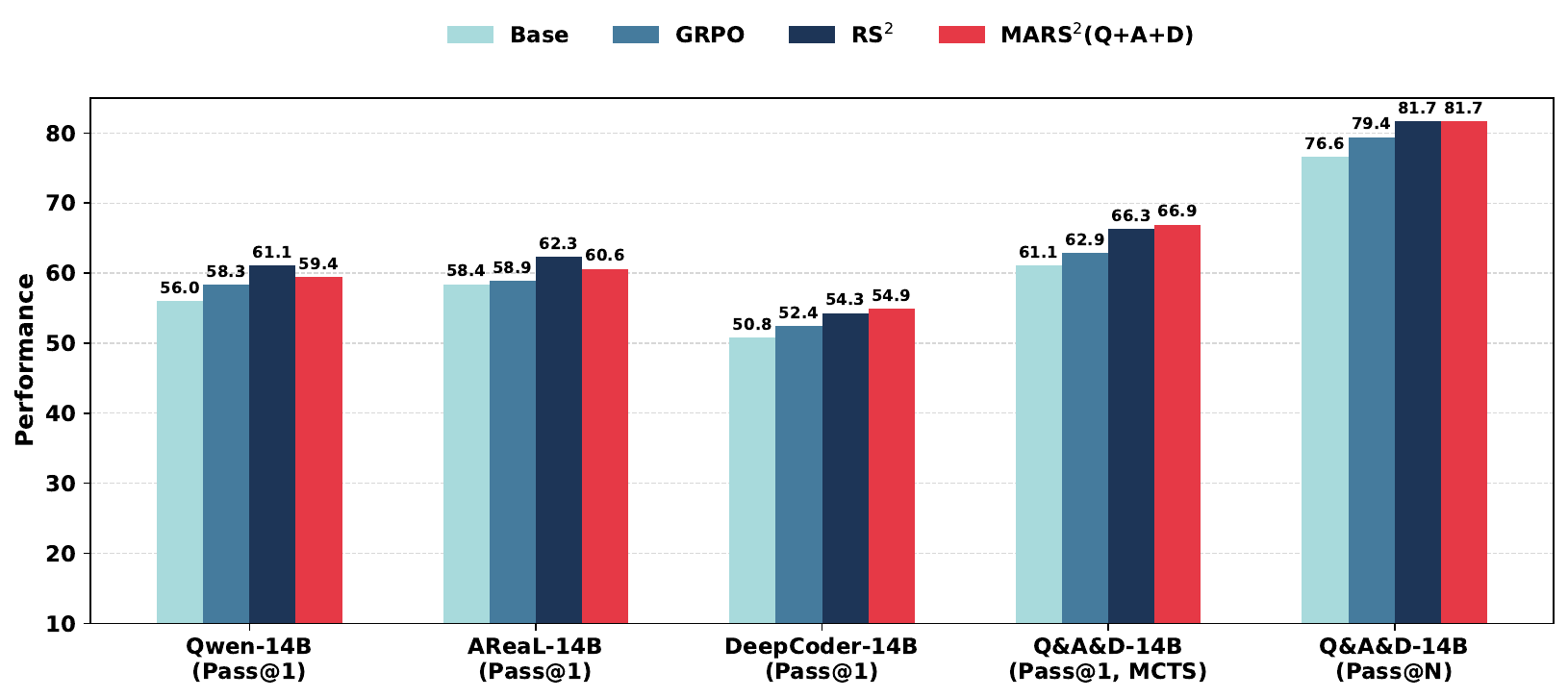}
  \vspace{-15pt}
  \caption{\textbf{Ablation on introducing a weaker agent in 14B-scale ensembles.}
  We augment the 14B two-agent setting (Qwen3-14B \& AReaL-14B) with a weaker model, DeepCoder-14B, to test robustness under imbalanced agent strength.}
  \vspace{-6pt}
  \label{fig:three14b}
\end{figure*}

 \subsubsection{Multi-Agent Reinforced Tree Search}
To address the inherent limitations of exploration imposed by single agent training, we construct multi-agent configurations at both the 8B and 14B model scales. Specifically, at the 8B scale, we consider combinations of Qwen3-8B and AReaL-8B, while at the 14B scale we evaluate combinations of Qwen3-14B and AReaL-14B. To ensure fair comparison, the total training data budget is strictly controlled and kept identical across all experimental settings.

\textbf{Individual performance and system-level collaboration.}
To examine how the number of agents affects individual model capability, we evaluate Pass@1 on five base LLMs, with results reported in Table~\ref{tab:main_results}. All agents trained under MARS$^2$ framework consistently outperform both their base counterparts and those trained with RS$^2$. For example, Qwen3-8B (58.3\%) achieves an absolute Pass@1 improvement of 8.0\% over the base model, 4.4\% over Vanilla GRPO, and a further 2.9\% gain beyond the peak performance attained by RS$^2$ training. AreaL-14B achieves a performance of 64.6\%, surpassing the 63.7\% of O4-Mini (Low).
The corresponding training curves in Figure~\ref{fig:matraincurve} further show that MARS$^2$ converges stably after an initial exploration phase, eventually surpassing Vanilla GRPO on all four paired models.
At the system level, as illustrated in Table~\ref{tab:main_results}, the heterogeneous system composed of Qwen3-8B and AReaL-8B achieves a Pass@1(MCTS) score of 61.7\%, representing a 4.5-point improvement over the base system (57.2\%), and clearly outperforming both Vanilla GRPO (56.0\%) and RS$^2$ (57.2\%). 
A similar trend is observed at the 14B scale. The Qwen3-14B \& AReaL-14B system trained with MARS$^2$ reaches a Pass@1(MCTS) score of 68.9\%, surpassing the corresponding base system (62.9\%) as well as Vanilla GRPO (61.7\%) and RS$^2$ (68.0\%). 
Moreover, consistent improvements in Pass@N are observed across both model scales, indicating that MARS$^2$ effectively enhances system-level collaboration and search efficiency under a fixed inference budget.

\subsection{Ablation Study}

\paragraph{Impact of Weaker Agents on MARS$^2$ Training.}


To further evaluate the robustness of MARS$^2$ under heterogeneous and imbalanced settings, we introduce \textbf{DeepCoder-14B} into the 14B-scale multi-agent configuration. Compared to Qwen3-14B and AReaL-14B, DeepCoder-14B exhibits substantially weaker single-model performance. This setting is designed to emulate realistic multi-model collaboration scenarios, where participating agents often possess uneven capabilities, and to assess whether MARS$^2$ remains effective under such conditions.
As shown in Figure~\ref{fig:three14b}, incorporating DeepCoder-14B leads to a noticeable reduction in the \textbf{Pass@1} gains of individual agents compared to the RS$^2$ setting. For instance, under RS$^2$, Qwen3-14B achieves a 5.1\% absolute improvement in Pass@1 over the base model, whereas this gain decreases to 3.4\% after introducing DeepCoder-14B. This phenomenon suggests that weaker agents may inject noisier trajectories into the shared search space 
weaker agents may inject noisier trajectories into the shared search space,
thereby partially interfering with fine-grained policy optimization at the individual level.

Nevertheless, from a system-level perspective, \textbf{Pass@1 (MCTS)}, which reflects collaborative search performance, continues to exhibit a monotonic improvement from Base to RS$^2$, and further to MARS$^2$. Meanwhile, \textbf{Pass@N} also shows consistent gains under this setting. These results indicate that although weaker agents may slightly constrain the individual performance ceiling, MARS$^2$ is still able to effectively aggregate complementary information across agents and maintain strong system-level advantages through tree-structured search and inter-agent feedback mechanisms.
Overall, these findings demonstrate that MARS$^2$ remains robust in heterogeneous multi-agent systems with uneven capability distributions, and that its system-level benefits do not rely on all participating agents having comparable single-model performance.

\definecolor{marsblue}{RGB}{220,235,252}
\begin{table*}[t]
\centering
\small
\setlength{\tabcolsep}{12pt}
\caption{\textbf{Diversity-oriented evaluation of different training paradigms on Qwen3-8B \& AReaL-8B system. MARS$^2$ demonstrates robust diversity performance, achieving the best average rank across different metrics.}}
\label{tab:q8b_results}
\begin{tabular}{c c c c c c c c} 
\toprule
\textbf{Method} & \textbf{Pass@N} & \textbf{AEC} & \textbf{DA@K} & \textbf{EA} & \textbf{NAUADC} & \textbf{G-Vendi} & \textbf{Avg.} \\
\midrule
Base           & 72.0(4) & 1.295(3) & 1.634(4) & 1.321(4) & 1.686(4) & \textbf{7.146(1)} & 3.3 \\
GRPO   & 72.0(4) & 1.190(4) & 1.645(3) & 1.333(3) & 1.707(3) & \underline{7.063}(2) & 3.2 \\
RS$^2$         & \underline{72.6}(2) & \underline{1.315}(2) & \underline{1.664}(2) & \underline{1.341}(2) & \underline{1.716}(2) & 6.403(4) & 2.3 \\
\rowcolor{marsblue} MARS$^2$       & \textbf{75.4(1)} & \textbf{1.431(1)} & \textbf{1.677(1)} & \textbf{1.348(1)} & \textbf{1.750(1)} & 6.901(3) & \textbf{1.3} \\
\bottomrule
\end{tabular}
\vspace{-5pt}
\end{table*}
\paragraph{Why Multi-Agent Tree Search Improves Reinforcement Learning: A Diversity-Centric Analysis.}
To better understand why multi-agent tree search effectively enhances training performance, we extend the conventional \textbf{Pass@N} metric by incorporating a set of fine-grained diversity measures, including AEC, DA@K, EA, NAUADC, and G-Vendi (Appendix~\ref{sec:appendix-evaluation setting} for detailed definitions). These metrics characterize exploration behavior from multiple perspectives, such as solution space coverage, structural dispersion, and distributional diversity.
We conduct a systematic comparison among Base, Vanilla GRPO, RS$^2$, and MARS$^2$ under an 8B-scale agent configuration. As reported in Table~\ref{tab:q8b_results}, MARS$^2$ consistently achieves the best average rank across different diversity metrics, with particularly pronounced advantages on AEC and DA@K, which capture global coverage of the solution space.
These results suggest that the effectiveness of multi-agent tree search does not primarily stem from repeatedly exploiting a small set of high-reward trajectories. Instead, by maintaining multiple agents with diverse policy biases, the search process continuously preserves a richer and more diverse candidate solution pool. The resulting increase in exploration diversity enables the training process to access a broader and more complementary set of high-quality trajectories, thereby providing more informative learning signals for policy optimization. This empirical evidence explains why MARS$^2$ is able to overcome the performance saturation observed in single-agent training.

\paragraph{Effect of Reward Shaping on Performance and Training Stability.}
\begin{figure}[t]
  \centering
  \vspace{-5pt}
  \includegraphics[width=\columnwidth]{pics/qwen_14b_reward_shaping.png}
  \caption{\textbf{Training curves of Qwen3-14B with and without reward shaping.} We report Pass@1 performance over training steps, showing that reward shaping leads to more stable optimization.}
  \vspace{-10pt}
  \label{fig:reward shaping}
\end{figure}
\vspace{-1em}
Finally, we examine the role of reward shaping in MARS$^2$ to better understand its impact on performance and training dynamics. Under otherwise identical settings, we conduct ablation experiments on Qwen3-14B with and without reward shaping, and report the training curves in Figure~\ref{fig:reward shaping}.
Without reward shaping, the model shows slow and unstable improvements in Pass@1 during early training, followed by a delayed performance jump around step 80. This behavior reflects a common limitation of single-policy optimization: exploration is initially dominated by the model’s own prior, resulting in sparse, high-variance reward signals and inefficient credit assignment along long reasoning trajectories. Consequently, effective policy updates are postponed until sufficiently informative trajectories are discovered.
By contrast, introducing reward shaping leads to a significantly smoother and more stable training process, with faster and more consistent performance gains. We attribute this effect to the fact that reward shaping provides denser, intermediate supervision aligned with the structured search process, effectively constraining degenerate behaviors (e.g., overlong or truncated outputs) and regularizing token-level discrepancies between rollout and training policies. These additional signals reduce gradient variance and improve the quality of early-stage updates, allowing the policy to better exploit informative search outcomes and converge to stronger solutions.
A further sensitivity analysis on the mixing coefficient $\lambda$ (Eq.~\ref{equ:reward baseline}) confirms that reward shaping plays a structural role rather than acting as a simple hyperparameter, with a moderate mix of parent and sibling signals yielding the best performance (Appendix~\ref{sec:appendix-Additional Performance}, Table~\ref{tab:lambda-sensitivity}).

\paragraph{Generalization to Mathematical Reasoning.}
To assess the generality of our framework beyond code generation, we additionally evaluate RS$^2$ on the MATH dataset, a widely-used benchmark for mathematical reasoning. Using Qwen2.5-3B-Instruct as the base model and a tree budget of 16 nodes,  RS$^2$ improves Pass@1 (MCTS) from 0.756 (Base) and 0.776 (GRPO) to \textbf{0.804}(Appendix \ref{sec:appendix-Additional Performance}, Table~\ref{tab:math-results}). This consistent gain demonstrates that the benefits of our tree-level credit assignment are not restricted to coding tasks, and transfer naturally to other structured reasoning domains.

\section{Conclusion}
\label{sec:conclusion}

We propose MARS$^2$, a reinforcement learning framework based on multi-agent systems that rethinks reasoning scaling through collaborative and tree-structured exploration.
By modeling the search tree as a learnable multi-agent environment, MARS$^2$ integrates policy interaction directly into the search process, enabling the exploration frontier to expand beyond the limitations of a single-policy prior.
Experiments across multiple model scales and configurations demonstrate that combining policy heterogeneity with structured tree search yields consistent improvements in both individual and system-level reasoning performance.
These results indicate that multi-agent tree search provides a scalable and robust alternative to single-agent, search-augmented reinforcement learning, and represents a promising direction for advancing reasoning systems.

\section*{Limitations}

This work primarily aims to establish the feasibility and effectiveness of integrating multi-agent collaboration with structured tree search in reinforcement learning.
Accordingly, our design emphasizes coordinated exploration and solution refinement, rather than optimizing for training efficiency.
In particular, multi-agent tree search relies on sequential interactions among agents, which may increase training time compared to single-agent approaches.This overhead stems from the reduced rollout parallelism caused by sequential tree expansion, rather than from any increase in data or compute consumption.
While such structure is crucial for enabling meaningful coordination and structured reasoning, developing more efficient search mechanisms that preserve these advantages remains an important direction for future research.

\section*{Acknowledgments}
This work was supported by Shanghai Artificial Intelligence Laboratory and supported by the National Natural Science Foundation of China (Grant No. 6250076080) and supported by the China Postdoctoral ScienceFoundation under Grant Number 2025M771537.

\bibliography{custom}

\begin{thebibliography}{36}
\providecommand{\natexlab}[1]{#1}

\bibitem[{An et~al.(2025)An, Xie, Li, Li, Zhang, Gong, Zhong, Xu, Qiu, Wang, and Kong}]{Polaris2025}
Chenxin An, Zhihui Xie, Xiaonan Li, Lei Li, Jun Zhang, Shansan Gong, Ming Zhong, Jingjing Xu, Xipeng Qiu, Mingxuan Wang, and Lingpeng Kong. 2025.
\newblock \href {https://hkunlp.github.io/blog/2025/Polaris} {Polaris: A post-training recipe for scaling reinforcement learning on advanced reasoning models}.

\bibitem[{Chern et~al.(2024)Chern, Fan, and Liu}]{chern2024combating}
Steffi Chern, Zhen Fan, and Andy Liu. 2024.
\newblock Combating adversarial attacks with multi-agent debate.
\newblock \emph{arXiv preprint arXiv:2401.05998}.

\bibitem[{Ester et~al.(1996)Ester, Kriegel, Sander, Xu et~al.}]{ester1996density}
Martin Ester, Hans-Peter Kriegel, J{\"o}rg Sander, Xiaowei Xu, and 1 others. 1996.
\newblock A density-based algorithm for discovering clusters in large spatial databases with noise.
\newblock In \emph{kdd}, volume~96, pages 226--231.

\bibitem[{Fu et~al.(2025)Fu, Gao, Shen, Zhu, Mei, He, Xu, Wei, Mei, Wang et~al.}]{areal}
Wei Fu, Jiaxuan Gao, Xujie Shen, Chen Zhu, Zhiyu Mei, Chuyi He, Shusheng Xu, Guo Wei, Jun Mei, Jiashu Wang, and 1 others. 2025.
\newblock Areal: A large-scale asynchronous reinforcement learning system for language reasoning.

\bibitem[{Guo et~al.(2025)Guo, Yang, Zhang, Song, Zhang, Xu, Zhu, Ma, Wang, Bi et~al.}]{guo2025deepseek}
Daya Guo, Dejian Yang, Haowei Zhang, Junxiao Song, Ruoyu Zhang, Runxin Xu, Qihao Zhu, Shirong Ma, Peiyi Wang, Xiao Bi, and 1 others. 2025.
\newblock Deepseek-r1: Incentivizing reasoning capability in llms via reinforcement learning.
\newblock \emph{arXiv preprint arXiv:2501.12948}.

\bibitem[{Hernandez{-}Leal et~al.(2017)Hernandez{-}Leal, Kaisers, Baarslag, and de~Cote}]{DBLP:journals/corr/Hernandez-LealK17}
Pablo Hernandez{-}Leal, Michael Kaisers, Tim Baarslag, and Enrique~Munoz de~Cote. 2017.
\newblock \href {https://arxiv.org/abs/1707.09183} {A survey of learning in multiagent environments: Dealing with non-stationarity}.
\newblock \emph{CoRR}, abs/1707.09183.

\bibitem[{Hou et~al.(2025)Hou, Hu, Li, Lu, Tang, and Dong}]{treerl}
Zhenyu Hou, Ziniu Hu, Yujiang Li, Rui Lu, Jie Tang, and Yuxiao Dong. 2025.
\newblock Treerl: Llm reinforcement learning with on-policy tree search.
\newblock In \emph{Proceedings of the 63rd Annual Meeting of the Association for Computational Linguistics (Volume 1: Long Papers)}, pages 12355--12369.

\bibitem[{Inoue et~al.(2025)Inoue, Misaki, Imajuku, Kuroki, Nakamura, and Akiba}]{inoue2025wider}
Yuichi Inoue, Kou Misaki, Yuki Imajuku, So~Kuroki, Taishi Nakamura, and Takuya Akiba. 2025.
\newblock Wider or deeper? scaling llm inference-time compute with adaptive branching tree search.
\newblock \emph{arXiv preprint arXiv:2503.04412}.

\bibitem[{Jain et~al.(2024)Jain, Han, Gu, Li, Yan, Zhang, Wang, Solar-Lezama, Sen, and Stoica}]{livecodebench}
Naman Jain, King Han, Alex Gu, Wen-Ding Li, Fanjia Yan, Tianjun Zhang, Sida Wang, Armando Solar-Lezama, Koushik Sen, and Ion Stoica. 2024.
\newblock Livecodebench: Holistic and contamination free evaluation of large language models for code.
\newblock \emph{arXiv preprint arXiv:2403.07974}.

\bibitem[{Jung et~al.(2025)Jung, Han, Lu, Hallinan, Acuna, Prabhumoye, Patwary, Shoeybi, Catanzaro, and Choi}]{jung2025prismatic}
Jaehun Jung, Seungju Han, Ximing Lu, Skyler Hallinan, David Acuna, Shrimai Prabhumoye, Mostafa Patwary, Mohammad Shoeybi, Bryan Catanzaro, and Yejin Choi. 2025.
\newblock Prismatic synthesis: Gradient-based data diversification boosts generalization in llm reasoning.
\newblock \emph{arXiv preprint arXiv:2505.20161}.

\bibitem[{Kryvosheieva et~al.(2025)Kryvosheieva, Sturua, Günther, Martens, and Xiao}]{kryvosheieva2025efficientcodeembeddingscode}
Daria Kryvosheieva, Saba Sturua, Michael Günther, Scott Martens, and Han Xiao. 2025.
\newblock \href {https://arxiv.org/abs/2508.21290} {Efficient code embeddings from code generation models}.
\newblock \emph{Preprint}, arXiv:2508.21290.

\bibitem[{Kwon et~al.(2023)Kwon, Li, Zhuang, Sheng, Zheng, Yu, Gonzalez, Zhang, and Stoica}]{kwon2023efficient}
Woosuk Kwon, Zhuohan Li, Siyuan Zhuang, Ying Sheng, Lianmin Zheng, Cody~Hao Yu, Joseph~E. Gonzalez, Hao Zhang, and Ion Stoica. 2023.
\newblock Efficient memory management for large language model serving with pagedattention.
\newblock In \emph{Proceedings of the ACM SIGOPS 29th Symposium on Operating Systems Principles}.

\bibitem[{Lee et~al.(2025)Lee, Chon, Jang, Lee, and Yu}]{lee-etal-2025-diversely}
Seonghyeon Lee, HeeJae Chon, Joonwon Jang, Dongha Lee, and Hwanjo Yu. 2025.
\newblock How diversely can language models solve problems? exploring the algorithmic diversity of model-generated code.
\newblock In \emph{Findings of the Association for Computational Linguistics: EMNLP 2025}, Suzhou, China. Association for Computational Linguistics.

\bibitem[{Li et~al.(2025)Li, Gu, Wen, Li, Xing, Guo, Zheng, Zhou, Qu, Zhou, Zhang, Shen, Liu, Lin, Yang, Zhang, and Huang}]{li2025treepo}
Yizhi Li, Qingshui Gu, Zhoufutu Wen, Ziniu Li, Tianshun Xing, Shuyue Guo, Tianyu Zheng, Xin Zhou, Xingwei Qu, Wangchunshu Zhou, Zheng Zhang, Wei Shen, Qian Liu, Chenghua Lin, Jian Yang, Ge~Zhang, and Wenhao Huang. 2025.
\newblock \href {https://arxiv.org/abs/2508.17445} {Treepo: Bridging the gap of policy optimization and efficacy and inference efficiency with heuristic tree-based modeling}.
\newblock \url{https://m-a-p.ai/TreePO}.
\newblock \emph{Preprint}, arXiv:2508.17445.

\bibitem[{Liu et~al.(2025)Liu, Du, Yang, Li, and Qiu}]{Marsrl2025}
Shulin Liu, Dong Du, Tao Yang, Yang Li, and Boyu Qiu. 2025.
\newblock Marsrl: Advancing multi-agent reasoning system via reinforcement learning with agentic pipeline parallelism.

\bibitem[{Lowe et~al.(2017)Lowe, Wu, Tamar, Harb, Abbeel, and Mordatch}]{DBLP:conf/nips/LoweWTHAM17}
Ryan Lowe, Yi~Wu, Aviv Tamar, Jean Harb, Pieter Abbeel, and Igor Mordatch. 2017.
\newblock \href {https://proceedings.neurips.cc/paper/2017/hash/68a9750337a418a86fe06c1991a1d64c-Abstract.html} {Multi-agent actor-critic for mixed cooperative-competitive environments}.
\newblock In \emph{Advances in Neural Information Processing Systems 30: Annual Conference on Neural Information Processing Systems 2017, December 4-9, 2017, Long Beach, CA, {USA}}, pages 6379--6390.

\bibitem[{Luo et~al.(2025)Luo, Tan, Huang, Patel, Ariyak, Wu, Shi, Xin, Cai, Weber et~al.}]{deepcoder2025}
Michael Luo, Sijun Tan, Roy Huang, Ameen Patel, Alpay Ariyak, Qingyang Wu, Xiaoxiang Shi, Rachel Xin, Colin Cai, Maurice Weber, and 1 others. 2025.
\newblock Deepcoder: A fully open-source 14b coder at o3-mini level.
\newblock \emph{Notion Blog}.

\bibitem[{Park et~al.(2025{\natexlab{a}})Park, Han, Guo, Ozdaglar, Zhang, and Kim}]{DBLP:conf/acl/Park0GOZK25}
Chanwoo Park, Seungju Han, Xingzhi Guo, Asuman~E. Ozdaglar, Kaiqing Zhang, and Joo{-}Kyung Kim. 2025{\natexlab{a}}.
\newblock \href {https://aclanthology.org/2025.acl-long.1459/} {Maporl: Multi-agent post-co-training for collaborative large language models with reinforcement learning}.
\newblock In \emph{Proceedings of the 63rd Annual Meeting of the Association for Computational Linguistics (Volume 1: Long Papers), {ACL} 2025, Vienna, Austria, July 27 - August 1, 2025}, pages 30215--30248. Association for Computational Linguistics.

\bibitem[{Park et~al.(2025{\natexlab{b}})Park, Han, Guo, Ozdaglar, Zhang, and Kim}]{park2025maporl}
Chanwoo Park, Seungju Han, Xingzhi Guo, Asuman~E Ozdaglar, Kaiqing Zhang, and Joo-Kyung Kim. 2025{\natexlab{b}}.
\newblock Maporl: Multi-agent post-co-training for collaborative large language models with reinforcement learning.
\newblock In \emph{Proceedings of the 63rd Annual Meeting of the Association for Computational Linguistics (Volume 1: Long Papers)}, pages 30215--30248.

\bibitem[{Shao et~al.(2024)Shao, Wang, Zhu, Xu, Song, Bi, Zhang, Zhang, Li, Wu et~al.}]{shao2024deepseekmath}
Zhihong Shao, Peiyi Wang, Qihao Zhu, Runxin Xu, Junxiao Song, Xiao Bi, Haowei Zhang, Mingchuan Zhang, YK~Li, Yang Wu, and 1 others. 2024.
\newblock Deepseekmath: Pushing the limits of mathematical reasoning in open language models.
\newblock \emph{arXiv preprint arXiv:2402.03300}.

\bibitem[{Shypula et~al.(2025)Shypula, Li, Zhang, Padmakumar, Yin, and Bastani}]{shypula2025evaluating}
Alexander Shypula, Shuo Li, Botong Zhang, Vishakh Padmakumar, Kayo Yin, and Osbert Bastani. 2025.
\newblock Evaluating the diversity and quality of llm generated content.
\newblock \emph{arXiv preprint arXiv:2504.12522}.

\bibitem[{Silver et~al.(2016)}]{silver2016alphago}
David Silver and 1 others. 2016.
\newblock Mastering the game of go with deep neural networks and tree search.
\newblock \emph{Nature}.

\bibitem[{Song et~al.(2025)Song, Kempe, and Munos}]{song2025outcomebased}
Yuda Song, Julia Kempe, and R{\'e}mi Munos. 2025.
\newblock \href {https://openreview.net/forum?id=VORSpYLBJ6} {Outcome-based exploration for {LLM} reasoning}.
\newblock In \emph{NeurIPS 2025 Workshop: Second Workshop on Aligning Reinforcement Learning Experimentalists and Theorists}.

\bibitem[{Thompson(1933)}]{Thompson1933ONTL}
William~R. Thompson. 1933.
\newblock \href {https://api.semanticscholar.org/CorpusID:120462794} {On the likelihood that one unknown probability exceeds another in view of the evidence of two samples}.
\newblock \emph{Biometrika}, 25:285--294.

\bibitem[{Yang et~al.(2025)Yang, Li, Yang, Zhang, Hui, Zheng, Yu, Gao, Huang, Lv et~al.}]{yang2025qwen3}
An~Yang, Anfeng Li, Baosong Yang, Beichen Zhang, Binyuan Hui, Bo~Zheng, Bowen Yu, Chang Gao, Chengen Huang, Chenxu Lv, and 1 others. 2025.
\newblock Qwen3 technical report.
\newblock \emph{arXiv preprint arXiv:2505.09388}.

\bibitem[{Yang et~al.(2024)Yang, Yang, Zhang, Hui, Zheng, Yu, Li, Liu, Huang, Wei, Lin, Yang, Tu, Zhang, Yang, Yang, Zhou, Lin, Dang, Lu, Bao, Yang, Yu, Li, Xue, Zhang, Zhu, Men, Lin, Li, Xia, Ren, Ren, Fan, Su, Zhang, Wan, Liu, Cui, Zhang, and Qiu}]{DBLP:journals/corr/abs-2412-15115}
An~Yang, Baosong Yang, Beichen Zhang, Binyuan Hui, Bo~Zheng, Bowen Yu, Chengyuan Li, Dayiheng Liu, Fei Huang, Haoran Wei, Huan Lin, Jian Yang, Jianhong Tu, Jianwei Zhang, Jianxin Yang, Jiaxi Yang, Jingren Zhou, Junyang Lin, Kai Dang, and 22 others. 2024.
\newblock \href {https://doi.org/10.48550/ARXIV.2412.15115} {Qwen2.5 technical report}.
\newblock \emph{CoRR}, abs/2412.15115.

\bibitem[{Yao et~al.(2025)Yao, Liu, Zhang, Dong, Shang, and Gao}]{yaoyour}
Feng Yao, Liyuan Liu, Dinghuai Zhang, Chengyu Dong, Jingbo Shang, and Jianfeng Gao. 2025.
\newblock Your efficient rl framework secretly brings you off-policy rl training, august 2025.
\newblock \emph{URL https://fengyao. notion. site/off-policy-rl}.

\bibitem[{Yu et~al.(2025)Yu, Zhang, Zhu, Yuan, Zuo, Yue, Dai, Fan, Liu, Liu et~al.}]{yu2025dapo}
Qiying Yu, Zheng Zhang, Ruofei Zhu, Yufeng Yuan, Xiaochen Zuo, Yu~Yue, Weinan Dai, Tiantian Fan, Gaohong Liu, Lingjun Liu, and 1 others. 2025.
\newblock Dapo: An open-source llm reinforcement learning system at scale.
\newblock \emph{arXiv preprint arXiv:2503.14476}.

\bibitem[{Zhang et~al.(2025{\natexlab{a}})Zhang, Liu, Zhu, Tian, Zeng, Jia, Fan, Lv, Zuo, Jiang, Liu, Wang, Wang, Zhao, Hua, Wang, Wang, Gao, Long, Sun, Ma, Cui, Bai, Ding, Qi, and Zhou}]{marti2025}
Kaiyan Zhang, Runze Liu, Xuekai Zhu, Kai Tian, Sihang Zeng, Guoli Jia, Yuchen Fan, Xingtai Lv, Yuxin Zuo, Che Jiang, Ziyang Liu, Jianyu Wang, Yuru Wang, Ruotong Zhao, Ermo Hua, Yibo Wang, Shijie Wang, Junqi Gao, Xinwei Long, and 7 others. 2025{\natexlab{a}}.
\newblock \href {https://github.com/TsinghuaC3I/MARTI} {Marti: A framework for multi-agent llm systems reinforced training and inference}.

\bibitem[{Zhang et~al.(2025{\natexlab{b}})Zhang, Lyu, Sun, Wang, Zhang, Hua, Wu, Guo, Wang, Muennighoff et~al.}]{zhang2025survey}
Qiyuan Zhang, Fuyuan Lyu, Zexu Sun, Lei Wang, Weixu Zhang, Wenyue Hua, Haolun Wu, Zhihan Guo, Yufei Wang, Niklas Muennighoff, and 1 others. 2025{\natexlab{b}}.
\newblock A survey on test-time scaling in large language models: What, how, where, and how well?
\newblock \emph{arXiv preprint arXiv:2503.24235}.

\bibitem[{Zhang and Xiong(2025)}]{DBLP:conf/acl/ZhangX25}
Shaowei Zhang and Deyi Xiong. 2025.
\newblock \href {https://aclanthology.org/2025.findings-acl.862/} {Debate4math: Multi-agent debate for fine-grained reasoning in math}.
\newblock In \emph{Findings of the Association for Computational Linguistics, {ACL} 2025, Vienna, Austria, July 27 - August 1, 2025}, pages 16810--16824. Association for Computational Linguistics.

\bibitem[{Zhao et~al.(2025)Zhao, Wu, Yue, Wu, Xu, Lin, Wang, Wu, Zheng, and Huang}]{zhao2025absolute}
Andrew Zhao, Yiran Wu, Yang Yue, Tong Wu, Quentin Xu, Matthieu Lin, Shenzhi Wang, Qingyun Wu, Zilong Zheng, and Gao Huang. 2025.
\newblock Absolute zero: Reinforced self-play reasoning with zero data.
\newblock \emph{arXiv preprint arXiv:2505.03335}.

\bibitem[{Zhao et~al.(2023)Zhao, Gu, Varma, Luo, Huang, Xu, Wright, Shojanazeri, Ott, Shleifer, Desmaison, Balioglu, Nguyen, Chauhan, Hao, and Li}]{DBLP:journals/corr/abs-2304-11277}
Yanli Zhao, Andrew Gu, Rohan Varma, Liang Luo, Chien{-}Chin Huang, Min Xu, Less Wright, Hamid Shojanazeri, Myle Ott, Sam Shleifer, Alban Desmaison, Can Balioglu, Bernard Nguyen, Geeta Chauhan, Yuchen Hao, and Shen Li. 2023.
\newblock \href {https://doi.org/10.48550/ARXIV.2304.11277} {Pytorch {FSDP:} experiences on scaling fully sharded data parallel}.
\newblock \emph{CoRR}, abs/2304.11277.

\bibitem[{Zheng et~al.(2025)Zheng, Liu, Li, Chen, Yu, Gao, Dang, Liu, Men, Yang et~al.}]{zheng2025group}
Chujie Zheng, Shixuan Liu, Mingze Li, Xiong-Hui Chen, Bowen Yu, Chang Gao, Kai Dang, Yuqiong Liu, Rui Men, An~Yang, and 1 others. 2025.
\newblock Group sequence policy optimization.
\newblock \emph{arXiv preprint arXiv:2507.18071}.

\bibitem[{Zhoubian et~al.(2025)Zhoubian, Zhang, and Tang}]{zhoubian2025restrlachievingaccuratecode}
Sining Zhoubian, Dan Zhang, and Jie Tang. 2025.
\newblock \href {https://arxiv.org/abs/2508.19576} {Rest-rl: Achieving accurate code reasoning of llms with optimized self-training and decoding}.
\newblock \emph{Preprint}, arXiv:2508.19576.

\bibitem[{Zuo et~al.(2025)Zuo, Zhang, Sheng, Qu, Cui, Zhu, Li, Zhang, Long, Hua et~al.}]{zuo2025ttrl}
Yuxin Zuo, Kaiyan Zhang, Li~Sheng, Shang Qu, Ganqu Cui, Xuekai Zhu, Haozhan Li, Yuchen Zhang, Xinwei Long, Ermo Hua, and 1 others. 2025.
\newblock Ttrl: Test-time reinforcement learning.
\newblock \emph{arXiv preprint arXiv:2504.16084}.

\end{thebibliography}
\clearpage
\twocolumn

\appendix

\raggedbottom

\section{Related Works}
\label{sec:related_work}
\paragraph{Reinforcement Learning for Reasoning.}
Reinforcement learning (RL)–based post-training has become a standard paradigm for improving the reasoning capabilities of large language models (LLMs). Representative approaches include PPO-style optimization and more recent group-based variants such as Group Relative Policy Optimization (GRPO)\citep{shao2024deepseekmath}, which have demonstrated strong performance on mathematical reasoning and code generation tasks. 
Notably, DeepSeek-R1\citep{guo2025deepseek} and its code-oriented extensions adopt GRPO-style training to improve single-sample accuracy via outcome-level feedback, while code-specialized models such as DeepCoder\citep{deepcoder2025} and AReaL\citep{areal} leverage RL-based fine-tuning to enhance execution correctness on programming benchmarks.
Despite these advances, prior work has observed that RL-based reasoning improvements are often constrained by the exploration behavior of a single policy\citep{song2025outcomebased}. Under the i.i.d. sampling assumption, exploration remains implicitly bounded by the model’s own prior, resulting in limited trajectory diversity and diminishing returns as training scales.

\paragraph{Search-Augmented Reinforcement Learning.}
To alleviate exploration inefficiency and reward sparsity in reinforcement learning, several works integrate explicit search mechanisms into the training process. Tree-structured approaches such as TreeRL~\citep{treerl}, REST-RL~\citep{zhoubian2025restrlachievingaccuratecode}, and TreePO~\citep{li2025treepo} incorporate Monte Carlo Tree Search (MCTS) to expand candidate reasoning trajectories and provide richer supervision signals, resulting in improved training stability and solution quality.
However, in most existing methods, the entire search tree is still driven by a single policy distribution. As a consequence, search behavior tends to increasingly concentrate on a small subset of high-probability branches as training progresses, limiting the effective expansion of the exploration frontier. Moreover, discrepancies between training-time search procedures and inference-time generation introduce a structural mismatch that restricts the transferability of search-induced behaviors, a limitation that has been discussed in prior work on search-based reinforcement learning~\citep{silver2016alphago}.

\paragraph{Multi-Agent Systems for LLM Reasoning.}
Multi-agent reinforcement learning (MARL) has been explored as a complementary approach to improve exploration through policy interactions\citep{DBLP:journals/corr/Hernandez-LealK17, DBLP:conf/nips/LoweWTHAM17}. In LLM reasoning, prior works such as MAPoRL\citep{park2025maporl} instantiate multi-agent collaboration via trajectory-level interactions, including multi-round debates\citep{DBLP:conf/acl/ZhangX25,chern2024combating}, self-play\citep{zhao2025absolute}, or voting-based protocols\citep{zuo2025ttrl}. Related systems such as MarsRL\citep{Marsrl2025} and MARTI\citep{marti2025} further study cooperative or competitive agent configurations to enhance robustness and reduce individual bias.
While effective in practice, these approaches typically organize agent interactions in flat or sequential structures, without explicitly coupling them to structured exploration mechanisms. Consequently, multi-agent collaboration remains largely decoupled from the search process itself, limiting its effectiveness for deep, multi-branch reasoning tasks that require systematic exploration and backtracking.

\section{RL Training Strategy}
\label{sec:appendix-training-strategy}

\paragraph{Asynchronous Updates under Stochastic Tree Search.}
During multi-agent tree search, each node expansion selects an agent according to an updatable Beta prior. 
Although the total rollout budget is fixed, the stochasticity of tree search induces mild variation in how many samples each agent receives per rollout.
Fully synchronous updates would therefore introduce unnecessary waiting and reduce throughput.
We instead adopt a buffer-based asynchronous update scheme\citep{areal, yu2025dapo, marti2025}: each agent maintains a local buffer and triggers a parameter update once its buffer reaches a preset threshold, without synchronizing with other agents.
In practice, we use relatively small search budgets during training for efficiency.
Under such low-budget settings, per-agent sample counts remain close, avoiding degenerate cases where some agents rarely update.

\paragraph{Stabilization Techniques and Fair Comparison.}
RLVR(Reinforcement Learning with Verifiable Rewards) for long-horizon reasoning can still be unstable in practice.
Following common practice in recent RL systems, we incorporate several stabilization techniques.
Importantly, we apply the same techniques to both MARS$^2$ and all baselines to ensure a fair comparison.

\begin{itemize}
    \item \textbf{GSPO.}
    We adopt GSPO~\cite{zheng2025group} to stabilize policy updates for long-chain outputs.
    GSPO replaces token-wise importance aggregation with a geometric mean over the whole sequence, reducing sensitivity to a few unstable tokens.

    \item \textbf{Overlong Penalty.}
    We observe that many low-quality trajectories are associated with truncated or excessively long generations.
    Following DAPO~\cite{yu2025dapo}, we apply a length-based reward shaping term $R_{\text{length}}(y)$.

    \smallskip
    \noindent
    \textbf{If $|y| \le L_{\text{max}} - L_{\text{cache}}$:}
    \begin{equation}
        R_{\text{length}}(y) = 0.
    \end{equation}

    \noindent
    \textbf{If $L_{\text{max}} - L_{\text{cache}} < |y| \le L_{\text{max}}$:}
    \begin{equation}
        R_{\text{length}}(y) = \frac{(L_{\text{max}} - L_{\text{cache}}) - |y|}{L_{\text{cache}}}.
    \end{equation}

    \noindent
    \textbf{If $|y| > L_{\text{max}}$:}
    \begin{equation}
        R_{\text{length}}(y) = -1.
    \end{equation}

    \item \textbf{TIS.}
    To mitigate instability caused by mismatches between inference-time rollouts and training-time distributions, we apply Token-wise Importance Sampling (TIS)~\cite{yaoyour}.
    Concretely, we align rollout probabilities produced by vLLM~\cite{kwon2023efficient} with the FSDP training distribution~\cite{DBLP:journals/corr/abs-2304-11277} via:
    \begin{equation}
    \mathrm{vllm\_kl}(j)
    =
    \log
    \frac{\pi_{\theta_j}^{\text{vllm}}(o_i \mid q)}
         {\pi_{\theta_j}^{\text{fsdp}}(o_i \mid q)}.
    \end{equation}
\end{itemize}

\noindent
Overall, these design choices improve training stability while avoiding method-specific advantages:
all stabilization techniques are enabled for both MARS$^2$ and baselines, so the gains primarily reflect the proposed multi-agent tree-search training framework.

\section{Evaluation Setting}
\label{sec:appendix-evaluation setting}


\subsection{Diversity Evaluation Setup.}

\label{sec:appendix-Diversity Setup}

We evaluate the diversity of code generation produced by baseline methods and our proposed approach on the \textbf{LiveCodeBench v6}~\citep{livecodebench} dataset.

Following the standard practice in diversity evaluation~\citep{shypula2025evaluating}, diversity metrics are designed to characterize structural variation within the valid solution space, rather than to mix successful and failed outputs. When success rates differ substantially across methods, directly computing diversity over all problems would conflate quality gaps with structural differences and reduce interpretability.

Accordingly, we retain only those problems for which all compared methods generate at least one functionally correct solution, obtaining a subset of 95 problems used consistently across all diversity evaluations. We emphasize that this filtering is not intended to exclude difficult problems, but rather to ensure that diversity metrics focus on structural variation within the valid solution space where the metrics are statistically meaningful.

We adopt the following models to support different aspects of diversity analysis:
\begin{itemize}
    \item \textbf{Semantic embeddings.} 
    We use \textbf{jina-code-embeddings-1.5b}~\citep{kryvosheieva2025efficientcodeembeddingscode} to obtain dense semantic representations of generated code solutions for embedding-based clustering metrics.
    
    \item \textbf{Algorithm equivalence judgment.} 
    To determine whether two functionally correct code solutions implement the same underlying algorithm, we employ \textbf{Qwen2.5-7B-Instruct}\citep{DBLP:journals/corr/abs-2412-15115} with temperature set to 0.
    This model is used to cluster solutions into algorithm-level groups required for computing algorithmic diversity metrics such as DA@K and EA.
    
    \item \textbf{Gradient-based diversity analysis.}
    For computing G-Vendi, we adopt \textbf{Qwen2.5-0.5B-Instruct}\citep{DBLP:journals/corr/abs-2412-15115} as a lightweight proxy model to extract loss gradients efficiently.
\end{itemize}

Unless otherwise specified, all diversity metrics are computed exclusively on functionally correct solutions to ensure that measured diversity corresponds to meaningful algorithmic or reasoning differences rather than superficial variations.

\subsection{Diversity Metrics}
\label{sec:appendix-Diversity Metrics}
\paragraph{Average Embedding Clusters (AEC).}
Embedding-based metrics have been widely used to assess semantic diversity in code generation.
For each problem, we encode all functionally correct solutions using \textbf{jina-code-embeddings-1.5b}~\citep{kryvosheieva2025efficientcodeembeddingscode}.
We then apply DBSCAN clustering~\cite{ester1996density} to group solutions based on semantic similarity.
We define the Average Embedding Clusters (AEC) as:
\begin{equation}
    \text{AEC} = \frac{1}{|Q|} \sum_{q \in Q} C(A_q),
\end{equation}
where $Q$ denotes the set of evaluated problems, $A_q$ is the set of correct solutions for problem $q$, and $C(A_q)$ is the number of embedding clusters obtained for that problem.
A higher AEC indicates that functionally correct solutions are more widely distributed in the semantic space, reflecting higher semantic diversity.

\paragraph{DA@K, EA, and NAUADC.}
Following~\cite{lee-etal-2025-diversely}, we adopt DA@K, EA, and NAUADC to quantify algorithmic diversity from complementary perspectives.
We first cluster functionally correct solutions into algorithm groups using \textbf{Qwen2.5-7B-Instruct}\citep{DBLP:journals/corr/abs-2412-15115}, which judges whether two solutions implement the same algorithm.
Let $N$ denote the total number of correct solutions for a problem, $M$ the number of algorithm clusters, and $s_m$ the size of the $m$-th cluster.
\textbf{DA@K} measures the expected number of distinct algorithms covered when sampling $K$ solutions:
\begin{equation}
DA@K = \sum_{m=1}^{M} \left( 1 - \frac{\binom{N - s_m}{K}}{\binom{N}{K}} \right).
\end{equation}
\textbf{EA} (Effective Algorithms) characterizes the entropy of the algorithm distribution:
\begin{equation}
EA = \exp\left(- \sum_{m=1}^{M} p_m \ln p_m \right),
\end{equation}
where $p_m = \frac{s_m}{N}$.
\textbf{NAUADC} integrates DA@K over varying sampling budgets to capture diversity robustness across solution set sizes:
\begin{equation}
    \text{NAUADC} = \frac{1}{K_{\max} - 1} \sum_{k=1}^{K_{\max}} \text{DA@}k.
\end{equation}
In our experiments, we set $K_{\max}=60$ to cover all correct solutions generated per problem.

\paragraph{G-Vendi.}
While the above metrics focus on code- and algorithm-level diversity, they do not explicitly capture diversity in reasoning strategies.
To address this, we adopt the G-Vendi metric~\cite{jung2025prismatic}, which measures diversity in the loss-gradient space of a proxy model.
Using \textbf{Qwen2.5-0.5B-Instruct} as a lightweight proxy, we compute the loss gradient for each sample $(x, y)$, where $x$ denotes the problem and $y$ includes both the chain-of-thought reasoning and the final answer:
\begin{equation}
    g_\theta(x, y) = -\nabla \log P(y \mid x; \theta).
\end{equation}
The gradients are normalized and projected into a lower-dimensional space via random projection.
We then construct a covariance matrix from the projected gradients and compute the Shannon entropy of its eigenvalue distribution.
Let $\{\lambda_i^K\}$ denote the normalized eigenvalues of this covariance matrix.
The G-Vendi score is defined as:
\begin{equation}
    \text{G-Vendi}(D) = \exp\left( -\sum_i \lambda_i^K \log \lambda_i^K \right).
\end{equation}
Higher G-Vendi values indicate greater diversity in cognitive strategies and reasoning behaviors across generated solutions.

\section{Additional Experiments Performance}
\label{sec:appendix-Additional Performance}
\begin{figure*}[t]
    \centering

    \begin{subfigure}[b]{0.24\textwidth}
        \centering
        \includegraphics[width=\linewidth]{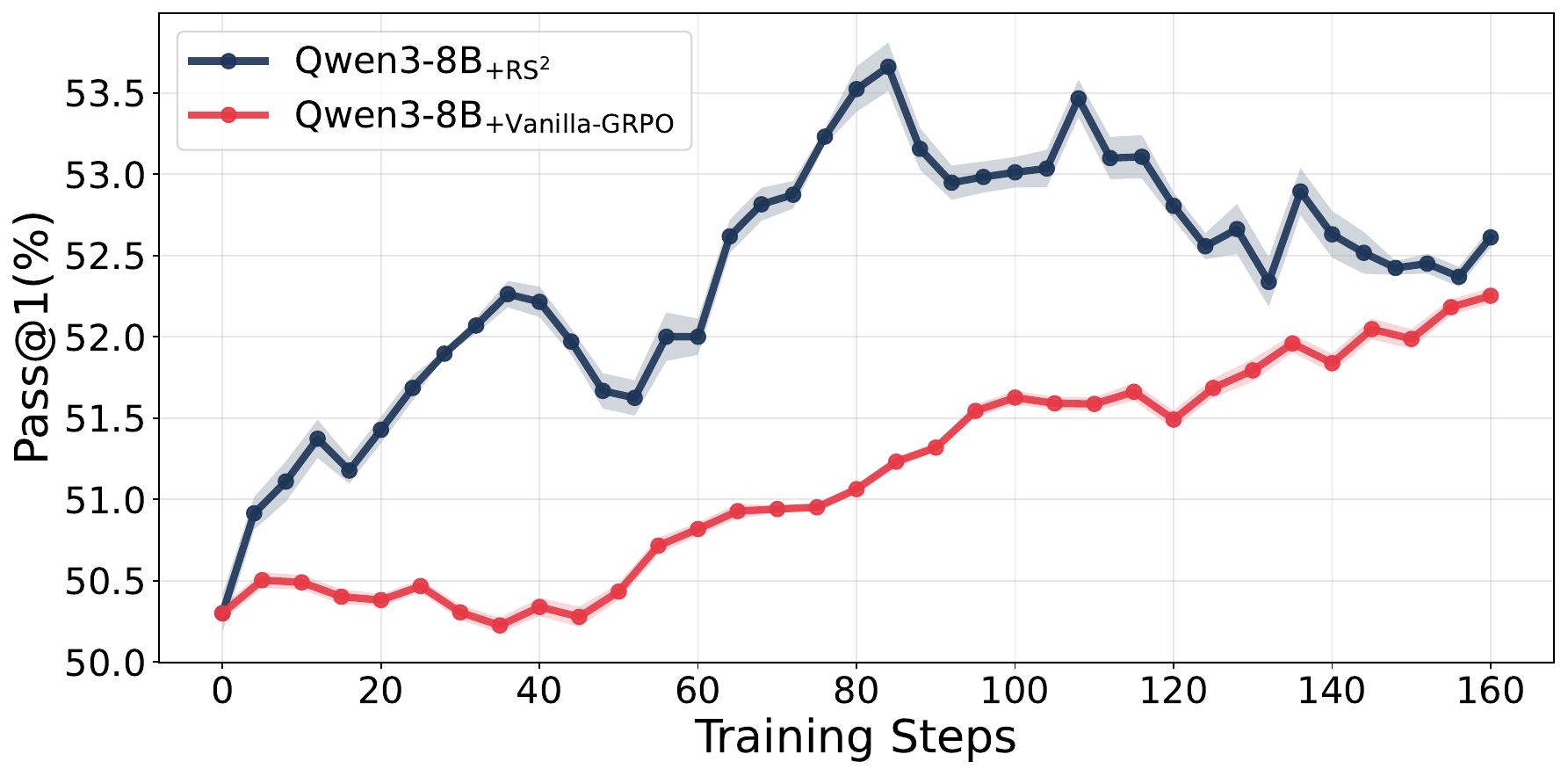}
        \subcaption{Pass@1 accuracy.}
        \label{fig:qwen3_8b_single_analysis}
    \end{subfigure}
    \hfill
    \begin{subfigure}[b]{0.24\textwidth}
        \centering
        \includegraphics[width=\linewidth]{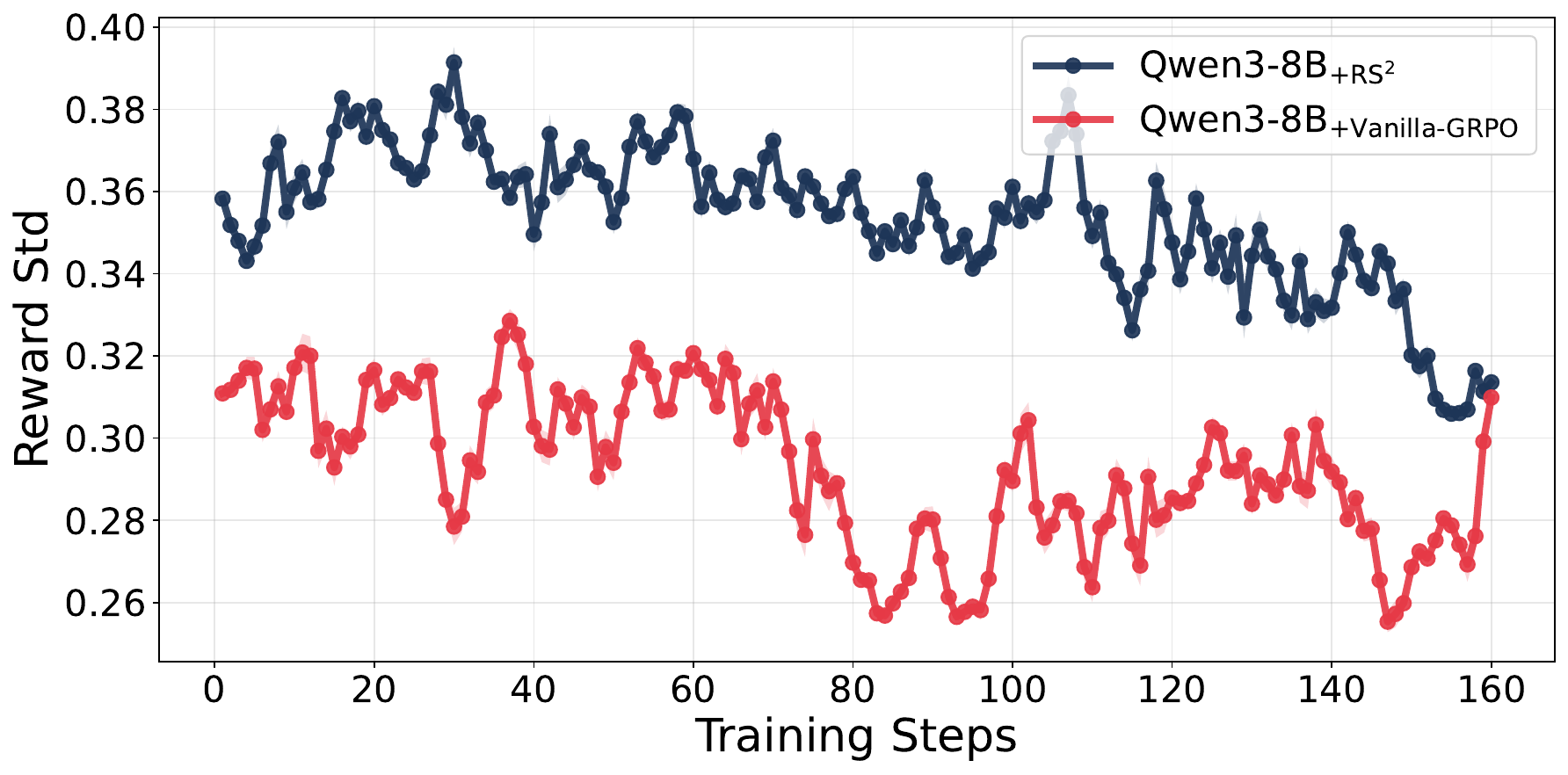}
        \subcaption{Reward standard deviation.}
        \label{fig:reward}
    \end{subfigure}
    \hfill
    \begin{subfigure}[b]{0.24\textwidth}
        \centering
        \includegraphics[width=\linewidth]{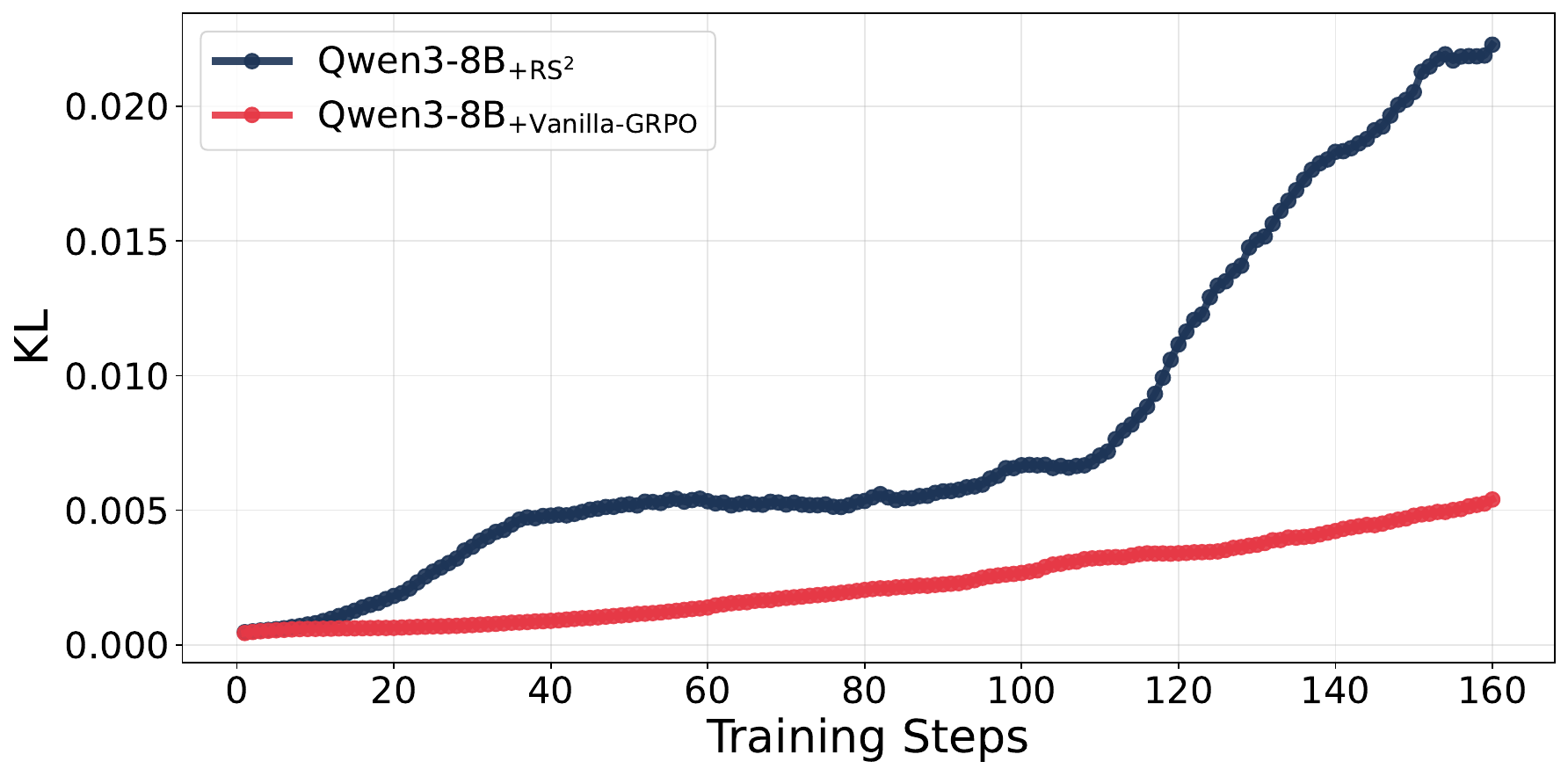}
        \subcaption{KL divergence.}
        \label{fig:kl}
    \end{subfigure}
    \hfill
    \begin{subfigure}[b]{0.24\textwidth}
        \centering
        \includegraphics[width=\linewidth]{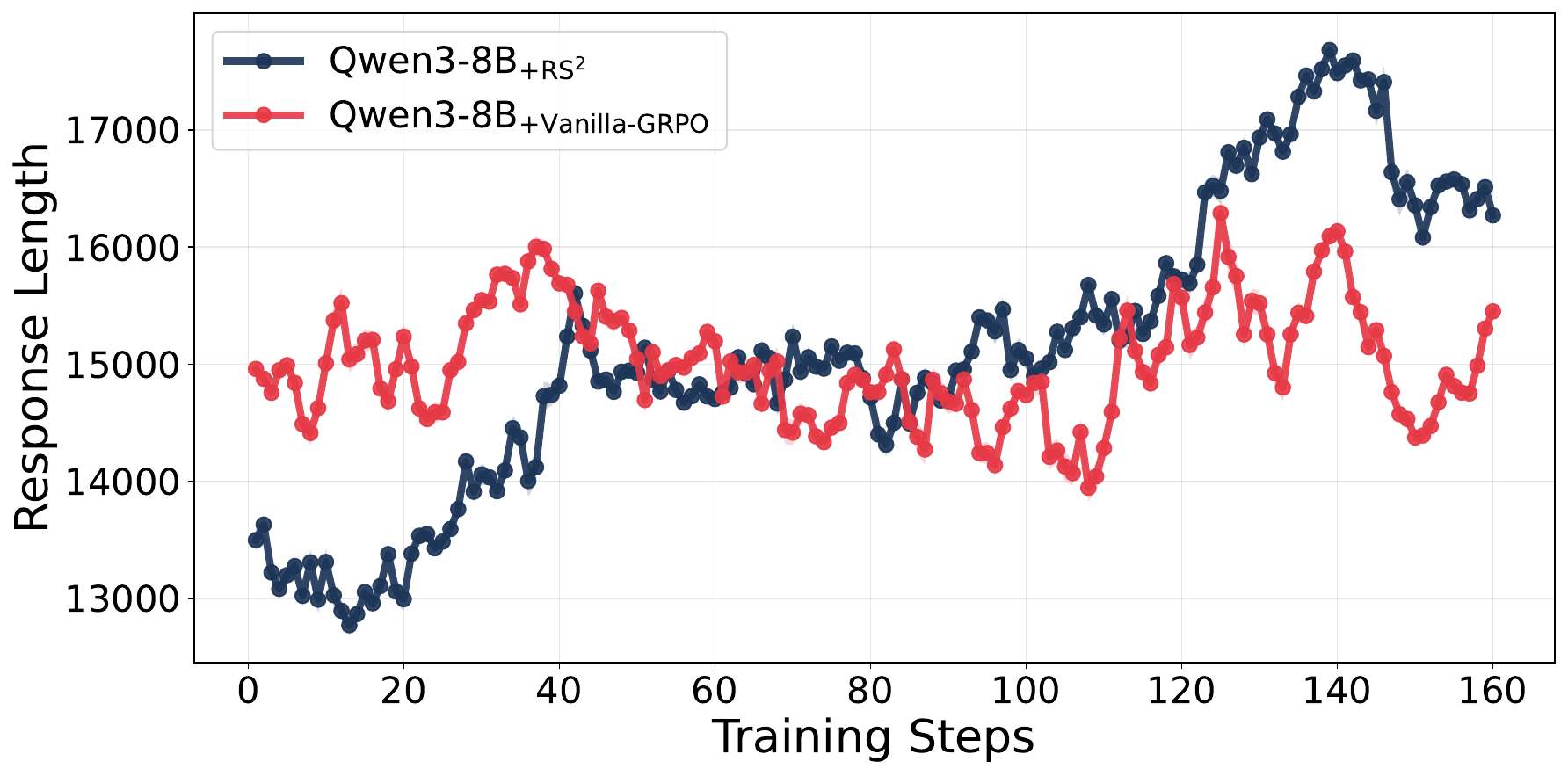}
        \subcaption{Response length.}
        \label{fig:response}
    \end{subfigure}
    
    \caption{Performance of RS$^2$ and GRPO on Qwen3-8B~\citep{yang2025qwen3} across training steps.}
    \vspace{-10pt}
    \label{fig:satraincurve}
\end{figure*}
\subsection{Performance Saturation in RS$^2$ Training.}
Despite the sustained improvements observed above, we further find that single-model RS$^2$ training exhibits a clear performance saturation effect(Figure~\ref{fig:satraincurve}). After an initial phase of rapid improvement, continued training yields diminishing returns: metrics such as Pass@1 plateau, and the variance of rewards associated with generated solutions gradually decreases.
This saturation phenomenon suggests that, although tree-structured search enhances exploration depth, the induced search distribution remains fundamentally constrained by the model’s own policy prior. As training progresses, the model increasingly revisits similar solution patterns, limiting its ability to discover novel and high-quality trajectories. In other words, tree-structured search alone is insufficient to fully overcome the intrinsic exploration boundaries imposed by a single policy distribution.

\subsection{Compute Budget}
\vspace{1em}

\begin{minipage}{0.48\textwidth}
\centering
\small
\captionof{table}{Training-time compute statistics for Qwen3-8B under GRPO and RS$^2$. Both methods use the same number of rollouts per sample and generate a comparable number of tokens, confirming that RS$^2$ does not benefit from a larger training or inference compute budget.}
\label{tab:compute-stats}
\begin{tabular}{lcc}
\toprule
\textbf{Method} & \textbf{Rollouts / Sample} & \textbf{Avg. Generated Tokens} \\
\midrule
GRPO   & 8 & 14{,}460 \\
RS$^2$ & 8 & 13{,}031 \\
\bottomrule
\end{tabular}
\end{minipage}

\vspace{0.5em}

\noindent
As shown in Table~\ref{tab:compute-stats}, RS$^2$ uses the same number of rollouts per sample as GRPO and in fact generates slightly fewer tokens on average. The reduced training throughput we occasionally observe under RS$^2$ and MARS$^2$ therefore does not stem from a larger data or compute budget; instead, it is primarily caused by the sequential nature of tree expansion, which lowers the parallel efficiency of rollout collection. The performance gains reported in Sec.~\ref{sec:experiments} can thus be safely attributed to the structured search and multi-agent collaboration rather than to additional training compute.

\subsection{Performance on MATH}
To probe whether the proposed tree-level credit assignment generalizes beyond executor-based reward signals, we additionally evaluate RS$^2$ on the MATH dataset, where node rewards $r_v$ are derived from answer matching rather than from code execution. Using Qwen2.5-3B-Instruct as the base model and a tree budget of 16 nodes, RS$^2$ improves Pass@1 (MCTS) from 0.756 (Base) and 0.776 (GRPO) to 0.804, as summarized in Table~\ref{tab:math-results}. The consistent gain suggests that the effectiveness of our tree-level credit assignment is not tied to executor-based rewards, and can be transferred to other reasoning tasks that supply node-level scalar evaluation signals.

\begin{table}[h]
\centering
\small
\caption{Pass@1 (MCTS, nodes=16) on the MATH dataset using Qwen2.5-3B-Instruct.}
\label{tab:math-results}
\begin{tabular}{lc}
\toprule
\textbf{Method} & \textbf{Pass@1 (MCTS)} \\
\midrule
Base   & 0.756 \\
GRPO   & 0.776 \\
RS$^2$ & \textbf{0.804} \\
\bottomrule
\end{tabular}
\vspace{-15pt}
\end{table}

\subsection{Latest-wins Selection Rule}
To verify that the "latest-wins" rule used in Pass@1 (MCTS) exploits the refinement structure of tree search rather than acting as an arbitrary heuristic, we compare it against random selection over the same candidate pool on Qwen3-4B-Instruct-2507. As shown in Table~\ref{tab:latest-wins}, latest-wins consistently outperforms random selection, indicating that deeper nodes in the search tree indeed tend to correspond to more refined and higher-quality solutions.

\begin{table}[h]
\centering
\small
\caption{Pass@1 (MCTS) of Qwen3-4B-Instruct-2507 (Base) on LiveCodeBench (v6) under different output selection strategies. Both strategies draw from the same candidate pool of nodes that pass the public tests; they differ only in which candidate is chosen as the final output.}
\label{tab:latest-wins}
\begin{tabular}{lc}
\toprule
\textbf{Selection Strategy} & \textbf{Pass@1 (MCTS)} \\
\midrule
Random select & 0.4171 \\
Latest-wins   & \textbf{0.4224} \\
\bottomrule
\end{tabular}
\vspace{-10pt}
\end{table}

\subsection{Stability of Training Search Budgets}
Our main experiments use a relatively small search budget during training for computational efficiency and controllable training stability. To verify that this choice does not mask instability at larger budgets, we additionally run RS$^2$ on Qwen3-4B-Instruct-2507 with tree-node budgets of 4, 8, and 16, and report the results in Table~\ref{tab:budget-scaling}. Increasing the budget from 4 to 8 nodes yields a clear improvement in Pass@1 (MCTS) (0.4079 $\to$ 0.4315), while further increasing to 16 nodes keeps the performance stable with no observable degradation or instability. Pass@1 remains nearly unchanged across all three budgets, indicating that single-model capability is not adversely affected. These results suggest that RS$^2$ scales gracefully with the training search budget, without the sample imbalance or non-stationarity issues that larger budgets might in principle introduce.

\begin{table}[h]
\centering
\small
\caption{Performance of RS$^2$ on Qwen3-4B-Instruct-2507 under different training-time tree-search budgets. Scaling the budget from 4 to 16 nodes produces stable improvements with no observable degradation or instability.}
\label{tab:budget-scaling}
\begin{tabular}{ccc}
\toprule
\textbf{Tree Nodes} & \textbf{Pass@1} & \textbf{Pass@1 (MCTS)} \\
\midrule
4  & 0.3886 & 0.4079 \\
8  & 0.3886 & \textbf{0.4315} \\
16 & \textbf{0.3943} & 0.4249 \\
\bottomrule
\end{tabular}
\vspace{-15pt}
\end{table}

\subsection{Sensitivity Analysis of Mixing Coefficient}
To further understand how reward shaping regulates hierarchical credit propagation, we conduct a sensitivity analysis of $\lambda$—the coefficient in Eq.~(\ref{equ:reward baseline}) that balances the parent-node reward against the sibling-node mean when forming the shaping baseline. Results on Qwen3-4B-Instruct-2507 are summarized in Table~\ref{tab:lambda-sensitivity}. We observe three consistent trends: (i) propagating parent rewards alone ($\lambda=0$) already yields a substantial improvement over the Base model (0.3400 $\to$ 0.4000 on Pass@1), showing that vertical signal propagation itself is effective; (ii) increasing $\lambda$ to $0.2$ and $0.4$ further improves performance, reaching the optimum at $\lambda=0.4$ (Pass@1 = 0.4571, Pass@1 (MCTS) = 0.4778); (iii) further increasing $\lambda$ to $0.8$ causes a clear performance drop, suggesting that over-reliance on sibling statistics dilutes discrimination among child nodes and weakens the exploration pressure toward promising branches. These results indicate that $\lambda$ plays a structural rather than cosmetic role: a moderate combination of parent rewards and sibling statistics provides the most consistent and stable credit signals, while either extreme degrades the hierarchical refinement dynamics.

\begin{table}[h]
\centering
\small
\caption{Sensitivity of reward shaping performance to the baseline mixing coefficient $\lambda$ on Qwen3-4B-Instruct-2507. A moderate value ($\lambda=0.4$) achieves the best trade-off between vertical (parent) and horizontal (sibling) credit signals.}
\label{tab:lambda-sensitivity}
\begin{tabular}{lcc}
\toprule
\textbf{Configuration} & \textbf{Pass@1} & \textbf{Pass@1 (MCTS)} \\
\midrule
w/o baseline           & 0.3400 & 0.4224 \\
$\lambda=0.0$  & 0.4000 & 0.4168 \\
$\lambda=0.2$  & 0.4286 & 0.4612 \\
$\lambda=0.4$  & \textbf{0.4571} & \textbf{0.4778} \\
$\lambda=0.8$  & 0.4400 & 0.4463 \\
\bottomrule
\end{tabular}
\vspace{-15pt}
\end{table}

\subsection{Effect of the Stabilization Suite}
To verify that the stabilization suite shared across all methods is not the dominant source of the gains reported in Sec.~\ref{sec:experiments}, we compare GRPO with and without the stabilization suite on Qwen3-8B under identical training steps (Table~\ref{tab:stabilization}). While stabilization does yield a meaningful improvement on GRPO ($49.43 \to 52.50$), the resulting performance is still substantially below RS$^2$ ($55.4$) and MARS$^2$ ($58.3$) on the same model (cf.\ Table~\ref{tab:main_results}). This confirms that the stabilization suite contributes to fair and reproducible training, but cannot independently account for the performance gains introduced by tree-level advantage and reward shaping.

\begin{table}[h]
\centering
\small
\caption{Effect of the stabilization suite on Qwen3-8B under Vanilla GRPO (Pass@1, measured at training step 100). Stabilization improves GRPO but does not reach the level of RS$^2$/MARS$^2$ in Table~\ref{tab:main_results}.}
\label{tab:stabilization}
\begin{tabular}{lc}
\toprule
\textbf{Method} & \textbf{Pass@1} \\
\midrule
GRPO (w/o stabilization) & 49.43 \\
GRPO (w/\ stabilization) & \textbf{52.50} \\
\bottomrule
\end{tabular}
\end{table}
\vspace{-10pt}

\section{Training Details}
\label{sec:appendix-Training Details}
\noindent
\renewcommand{\arraystretch}{0.9}
\begin{minipage}{0.5\textwidth}  
\centering
\small
\captionof{table}{Training configurations and hyperparameters for MARS$^2$.}
\vspace{-5pt}
\begin{tabularx}{\textwidth}{lX}
\toprule
\textbf{Parameter} & \textbf{Value} \\
\midrule

\multicolumn{2}{l}{\textbf{Model and Workflow Setup}} \\
MCTS Nodes & 8 / 16 \\
Max Prompt Length & 4096 \\
Max Generation Length & 32768 \\
Eval Generation Length & 32768 \\
Max Sequence Length & 40000 \\
Overlong Buffer Length & 2048 \\

\midrule
\multicolumn{2}{l}{\textbf{Cluster Configuration}} \\
Reference Model & 8 GPUs per model \\
Actor model & 8 GPUs per model \\
vLLM Engines & 8 per model \\
Tensor Parallel Size & 1 \\
vLLM GPU Memory Utilization & 0.85 \\

\midrule
\multicolumn{2}{l}{\textbf{Training Hyperparameters}} \\
Temperature & 1.0 \\
Training Batch Size & 256 \\
Micro Train Batch Size & 1 \\
Rollout Batch Size & 512 \\
Micro Rollout Batch Size & 1 \\
Samples per Prompt & 1 \\
Max Epochs & 1 \\
Random Seed & 42 \\
Learning Rate (Actor) & $1 \times 10^{-6}$ \\
Discount Factor $\gamma$ & 1.0 \\
ZeRO Stage & 3 \\
Precision & bfloat16 \\

\midrule
\multicolumn{2}{l}{\textbf{RL and PPO Settings}} \\
KL Loss & Enabled \\
Initial KL Coefficient & $1 \times 10^{-3}$ \\
KL Estimator & K3 \\
Reward Normalization & Enabled \\
Sample Packing & Enabled \\
Dynamic Reward Filter Range & [0, 1] \\
Importance Sampling Correction (vLLM) & Enabled \\
IS Truncation Threshold & 2 \\

\midrule
\multicolumn{2}{l}{\textbf{Inference Settings}} \\
Top-$p$ & 1.0 \\
Evaluation Temperature & 1.0 \\

\bottomrule
\end{tabularx}
\label{tab:training-config}
\end{minipage}


\section{Prompt Template}
\label{sec:appendix-Prompt Template}
\begin{tcolorbox}[title=Wrong-Answer Feedback Prompt]
\label{box:ab-mcts-prompt}
You are an expert Python programmer. You will be given a question (problem specification) and will generate a correct Python program that matches the specification and passes all tests. Here's your last question and your answer: 

Question:\{\}

Format: Read the inputs from stdin solve the problem and write the answer to stdout (do not directly test on the sample inputs). Enclose your code within delimiters as follows. Ensure that when the python program runs, it reads the inputs, runs the algorithm and writes output to STDOUT.

Your previous code:\{\}

Result: Wrong

Summary

Your solution is correct for 0 problems!, please re-implement your code. In addition to passing the test cases, try to ensure that your code can handle various input cases as much as possible, including edge cases and exceptions.

Answer: (use the provided format with backticks)
\end{tcolorbox}

\begin{tcolorbox}[title=Correct-Answer Feedback Prompt]
\label{box:ab-mcts-prompt}
You are an expert Python programmer. You will be given a question (problem specification) and will generate a correct Python program that matches the specification and passes all tests. Here's your last question and your answer: 

Question:\{\}

Format: Read the inputs from stdin solve the problem and write the answer to stdout (do not directly test on the sample inputs). Enclose your code within delimiters as follows. Ensure that when the python program runs, it reads the inputs, runs the algorithm and writes output to STDOUT.

Your previous code:\{\}

Result: Correct

Summary

Your solution is correct for all the problems!
Next, to ensure the robustness of your code, please make sure it can handle all possible input cases, including edge cases and exceptions.

Answer: (use the provided format with backticks)
\end{tcolorbox}


\end{document}